\documentclass[11pt,a4paper]{article}
\usepackage[left=2cm,right=2cm,top=2cm,bottom=2cm]{geometry}

\usepackage{times}
\usepackage{mathptmx}

\usepackage{amsmath}
\usepackage{amsfonts}
\usepackage{amssymb}

\author{Motaz Saad and David Langlois and Kamel Sma\"{i}li}
\title{Building and Aligning Comparable Corpora}
\date{}

\usepackage{graphicx} 
\usepackage{booktabs} 
\usepackage{adjustbox}

\usepackage[utf8]{inputenc}
\usepackage{amsmath}
\usepackage{amsfonts}
\usepackage{amssymb}
\usepackage{graphicx}

\usepackage[linesnumbered,boxruled,vlined]{algorithm2e}
\usepackage{hyperref}
\hypersetup{colorlinks=false}
\usepackage{etex} 
\usepackage{arabtex}
\usepackage{scalefnt}
\usepackage{utf8}
\setcode{utf8}
\usepackage{color}
\usepackage{eurosym}
\usepackage{url}
\usepackage{pbox}
\usepackage{multirow}
\usepackage{relsize}		
\usepackage{wrapfig}
\usepackage{minibox}
\usepackage{bchart}
\usepackage{wasysym}

\everymath{\displaystyle}
\usepackage{pgfplots}
\pgfplotsset{compat=newest}
\usepackage{tikz,lipsum}
\usetikzlibrary{matrix,decorations.pathreplacing,positioning}
\usetikzlibrary{shapes,shadows,arrows,patterns}
\tikzset{    
    multidocs/.style={double copy shadow={shadow xshift=-0.4ex,shadow yshift=0.4ex},fill=white,draw=black},
}

\SetCommentSty{mycommfont}
\usepackage{xcolor} 
\def\HiLi{\leavevmode\rlap{\hbox to \hsize{\color{yellow!50}\leaders\hrule height .8\baselineskip depth .5ex\hfill}}}

\DeclareFontFamily{U}{xnsh}{}
\DeclareFontShape{U}{xnsh}{m}{n}{                                  
   <-6> sfixed * [3.0] xnsh14
      <6-10> s * [1.2] xnsh14
         <10><10.95><12><14.4><17.28><20.74><24.88> s * [1.2] xnsh14
         }{}
\DeclareFontShape{U}{xnsh}{bx}{n}{
   <-6> sfixed * [3.0] xnsh14bf
   <6-10> s * [1.2] xnsh14bf
   <10><10.95><12><14.4><17.28><20.74><24.88> s * [1.2] xnsh14bf
}{}

\begin{document}
\sloppy 

\tikzstyle{decision} = [diamond, draw]
\tikzstyle{line} = [draw]
\tikzstyle{arrow} = [draw,>=triangle 45,->]
\tikzstyle{dash} = [draw, dashed]
\tikzstyle{elli}=[ellipse, draw]
\tikzstyle{rec} = [rectangle, draw]
\tikzstyle{decision} = [diamond, draw]
\tikzstyle{round} = [circle, draw]
\tikzstyle{corpus} = [shape=document, multidocs, draw]
\tikzstyle{doc} = [shape=document, draw, minimum width=0.7cm] 

\makeatletter
\pgfdeclareshape{document}{
\inheritsavedanchors[from=rectangle] 
\inheritanchorborder[from=rectangle]
\inheritanchor[from=rectangle]{center}
\inheritanchor[from=rectangle]{north}
\inheritanchor[from=rectangle]{south}
\inheritanchor[from=rectangle]{west}
\inheritanchor[from=rectangle]{east}
\backgroundpath{
\southwest \pgf@xa=\pgf@x \pgf@ya=\pgf@y
\northeast \pgf@xb=\pgf@x \pgf@yb=\pgf@y
\pgf@xc=\pgf@xb \advance\pgf@xc by-10pt 
\pgf@yc=\pgf@yb \advance\pgf@yc by-10pt
\pgfpathmoveto{\pgfpoint{\pgf@xa}{\pgf@ya}}
\pgfpathlineto{\pgfpoint{\pgf@xa}{\pgf@yb}}
\pgfpathlineto{\pgfpoint{\pgf@xc}{\pgf@yb}}
\pgfpathlineto{\pgfpoint{\pgf@xb}{\pgf@yc}}
\pgfpathlineto{\pgfpoint{\pgf@xb}{\pgf@ya}}
\pgfpathclose
\pgfpathmoveto{\pgfpoint{\pgf@xc}{\pgf@yb}}
\pgfpathlineto{\pgfpoint{\pgf@xc}{\pgf@yc}}
\pgfpathlineto{\pgfpoint{\pgf@xb}{\pgf@yc}}
\pgfpathlineto{\pgfpoint{\pgf@xc}{\pgf@yc}}
}
}
\makeatother

\maketitle

\begin{abstract}  

Comparable corpus is a set of topic aligned documents in multiple languages, which are not necessarily translations of each other. These documents are useful for multilingual natural language processing when there is no parallel text available in some domains or languages. In addition, comparable documents are informative because they can tell what is being said about a topic in different languages. In this paper, we present a method to build comparable corpora from Wikipedia encyclopedia and EURONEWS website in English, French and Arabic languages. We further experiment a method to automatically align comparable documents using cross-lingual similarity measures. We investigate two cross-lingual similarity measures to align comparable documents. The first measure is based on  bilingual dictionary, and the second measure is based on Latent Semantic Indexing (LSI). Experiments on several corpora show that the Cross-Lingual LSI (CL-LSI) measure outperforms the dictionary based measure. Finally, we collect English and Arabic news documents from the British Broadcast Corporation (BBC) and from ALJAZEERA (JSC) news website respectively. Then we use the CL-LSI similarity measure to automatically align comparable documents of BBC and JSC. The evaluation of the alignment shows that CL-LSI is not only able to align cross-lingual documents at the topic level, but also it is able to do this at the event level.

\end{abstract}


\section{Introduction}

Multilingual texts (parallel or comparable) are useful in several NLP applications such as bilingual lexicon extraction \cite{Li2010}, cross-lingual information retrieval \cite{Knoth2011} and machine translation \cite{Delpech2011}.  A parallel corpus is a collection of aligned sentences, which are translations of each other. Parallel corpora are acquired using human translators, but this is time consuming and requires a lot of human being efforts.

Comparable corpora can be obtained easily from the web. The emergence of Web 2.0 technologies enlarged web contents in many languages. Newspaper websites and Encyclopedias are ideal for collecting comparable documents. But aligning these texts is a challenging task.

\begin{figure}[!htb]
\begin{center}
\fbox{
\includegraphics[width=\linewidth,height=\textheight,keepaspectratio]{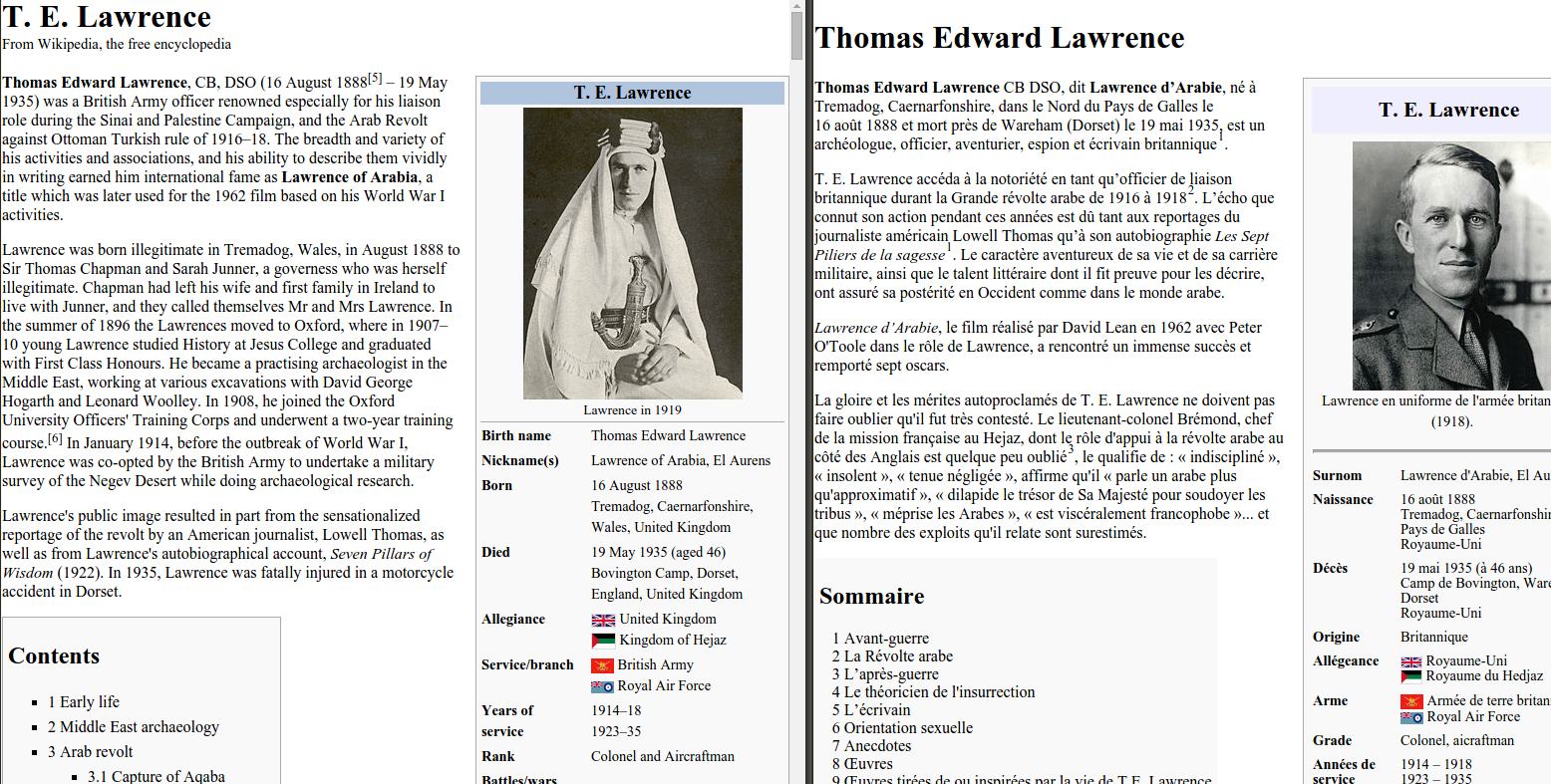}
}
\end{center}
\caption{English-French comparable documents from Wikipedia}
\label{fig:wikipeida_comparable_lowarence}
\end{figure}

\begin{figure}[!htb]
\begin{center}
\fbox{
\includegraphics[width=\linewidth,height=\textheight,keepaspectratio]{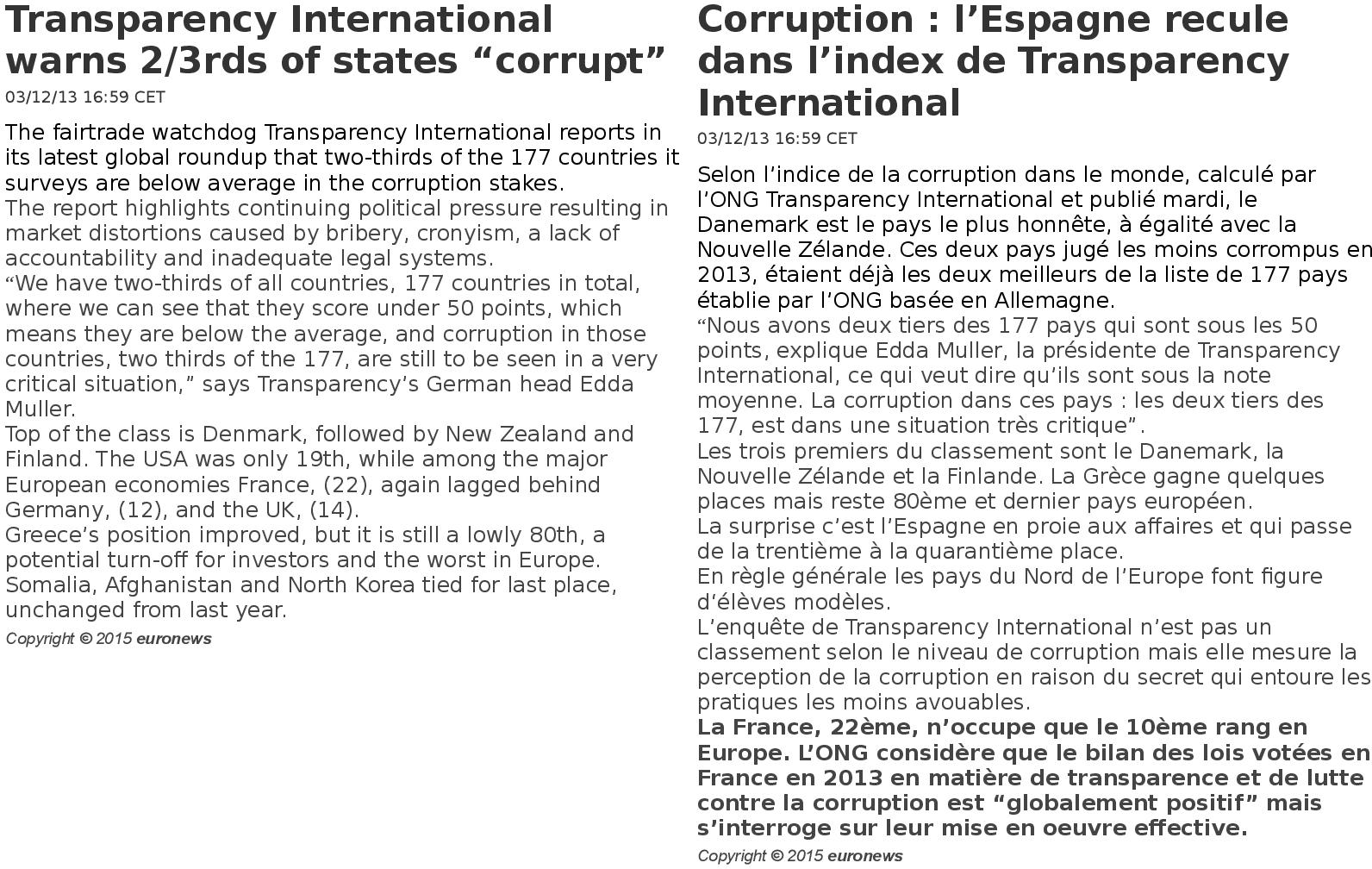}
}
\end{center}
\caption{English-French comparable news article from EURONEWS}
\label{fig:enews_comparable_corruption}
\end{figure}

Figure  \ref{fig:wikipeida_comparable_lowarence} shows an example of English-French Wikipedia comparable documents related to a biography of a person. It can be noted from the figure that the first paragraph in the English document is longer than the French one, it also provides more information about the person. It can be also noted that the English and the French texts of these comparable documents have different views to this person. Another example of comparable documents is the news articles from EURONEWS shown in Figure \ref{fig:enews_comparable_corruption}. The news article is related to a report about transparency and corruption in the world. Despite the documents are related to the same news story, the translations of their titles are different. Furthermore, the first paragraph in the English and in the French documents are different. The French document also has an additional paragraph that describes the position of France in this report. This paragraph does not exist in the English document.

Comparable corpora can be a good alternative to parallel corpora, when this resource is not available or not enough for specific domains or languages.

It can be noted from comparable document examples presented in Figures \ref{fig:wikipeida_comparable_lowarence} and \ref{fig:enews_comparable_corruption}, that comparable documents are informative when one is interested in what is being said about a topic in the other languages. So retrieving comparable documents can be useful in many applications beyond enriching language resources. For example, a journalist may be interested in what is being said about a news event covered in local and foreign news agencies, or a customer may be interested in exploring a product reviews written in different languages.

In this paper, we present in Section \ref{sec:corpora} a method to build comparable corpora from Wikipedia encyclopedia\footnote{www.wikipedia.org} and EURONEWS\footnote{www.euronews.com} websites In Arabic, French, and English languages \cite{Saad2013cilc}. Such resources does not exist publicly for researchers, so they are important and useful for the research community. We put the Wikipedia corpus online for research purposes\footnote{\url{http://sourceforge.net/projects/crlcl/}}. Then in Section \ref{sec:similarity_measure}, we present two cross-lingual similarity measures \cite{Saad2014poltal} that are used in this work for two tasks: (1) document pair retrieval, and (2) aligning comparable documents (Section \ref{sec:cross_lingual_alignment}). In the first task, the corpus is already aligned and the objective is to retrieve the target document by providing the source document as a query. In this task we apply the similarity measures on several corpora and we compare their performance, then we choose the best measure of the two measures to achieve the second task. In the second task, the corpus is not aligned, and the objective is to automatically align comparable source and target documents which are related to the same context. The resultant English-Arabic comparable news documents do not exist in research community, so our corpus is interesting and very useful for other researchers and they open new research directions.

\section{Related Words}
\label{sec:related_work}

\subsection{Comparable Corpora}

Non-parallel corpora can have different levels of comparability. In \cite{Fung2004}, the authors proposed three levels for non-parallel corpora. These levels are \textit{noisy-parallel}, \textit{comparable} and \textit{quasi-comparable} corpora. Texts in the \textit{noisy-parallel} corpus have many parallel sentences roughly in the same order. Texts in the \textit{comparable} corpus have topic aligned documents, which are not necessarily translations of each other. As for the \textit{quasi-comparable} corpus, it has bilingual documents that are not necessarily related to the same topic.

Most of researchers are interested in comparable corpora because they can be used  to extract parallel texts for different purposes. This interest continues to receive increasing attention from the research community. For example, ACCURAT\footnote{\url{www.accurat-project.eu}} project \cite{ACCURAT2012,ACCURAT2012b} is a research project dedicated to find methods and techniques to overcome the problem of lacking linguistic resources for under-resourced languages and narrow domains. The ultimate objective is to exploit comparable data to improve the quality of machine translation for such languages. The ACCURAT project researches for methods to measure comparable corpora and use them to achieve the project's objective. The project also research methods for alignment and extraction of lexical resources. Other researchers also considered comparable corpora to improve the quality of machine translation such in \cite{Smith2010,AbdulRauf2011} and to extract bilingual lexicons such in \cite{Li2010}.

Many researchers collected comparable corpora in many languages. For instance, authors of \cite{Barron2015} extracted Spanish-English comparable corpus, \cite{CHU2014} constructed a Chinese-Japanese corpus, \cite{Otero2009} collected Spanish-Portuguese-English corpus, and \cite{Ion2010} collected English-Romanian corpus. In this paper we build Arabic-French-English comparable corpora, and we make it available publicly for the research community. Such resources are important and useful because they are not available freely for the research community.

In the next section, we review some cross-lingual similarity measures, which are proposed to measure comparable corpora.

\subsection{Cross-lingual Similarity Measures}

A document similarity measure is a function that quantifies the similarity between the content of two documents \cite{IRbook}. This function gives a numerical score to quantify the similarity. In data mining and machine learning, Euclidean, Manhattan or Mahalanobis metrics are usually used to measure the similarity of two objects \cite{Witten2011}, while in text mining and information retrieval, cosine similarity or Jaccard index is used.

To measure the similarity between two text documents, they must be represented as vectors. To produce a document vector, the text document is transformed into a Bag Of Words (BOW), i.e., the text document is treated as a set of unstructured words. Representing a collection of documents as BOW is called Vector Space Model (VSM) or term-document matrix.

A cross-lingual similarity measure is a special case of document similarity measure, where the source and target documents are written in different languages. The cross-lingual measure can be used to identify a pair of comparable documents, i.e., to retrieve a target document for a given source document. Many methods have been proposed to measure the similarity of comparable documents. These methods are based on bilingual dictionaries, on Cross-Lingual Information Retrieval (CL-IR), or based on Cross-Lingual Latent Semantic Indexing (CL-LSI). These methods are described in the next sections.

\subsection{Dictionary-based Methods}

In dictionary-based methods, two cross-lingual documents $d_s$ and $d_t$ are comparable if most of words in $d_s$ are translations of words in $d_t$. A bilingual dictionary can be used to look-up the translations of words in both documents.

Matched words in source and target documents can be weighted using binary or $\mathit{tfidf}$ weighting schemes. The similarity can be measured on binary terms (1 $\rightarrow$ term present, 0 $\rightarrow$ term absent) as follows \cite{Li2010,Otero2011}:

\begin{equation}
sim(d_s,d_t) = \frac
{\vert w_s \rightarrow w_t \vert + \vert w_t \rightarrow w_s \vert}
{\vert d_s \vert + \vert d_t \vert }
\end{equation}

where $\vert w_s \rightarrow w_t \vert$ is the number of source words ($w_s$) that have translations in the target document ($d_t$), and  $\vert w_t \rightarrow w_s \vert$ is the number of target words ($w_t$) that have translations in the source document ($d_s$).  $\vert d_s \vert + \vert d_t \vert$ is the size (number of words) of the source and the target documents. 

The $\mathit{tfidf}$ weighting scheme reflects how a term is important to a document in a corpus. The $\mathit{tfidf}$ increases the weight to words that appear frequently in the document, but this increase is controlled by the frequency of the word in the corpus (document frequency). Thus, common words, that appeared in the most of documents, get less weight (less discriminative) than words that appeared in some documents (more discriminative). Cosine similarity measure is usually used to measure the similarity between vectors weighted with $\mathit{tfidf}$. In cross-lingual documents, the source and target document vectors are generated from the matched words (translation of each others) between source and target documents. Then the $\mathit{tfidf}$ is computed for source and target entries. 

Several similarity measures based on bilingual dictionary have been proposed. Some of them consider the similarity at the corpus level \cite{Li2010}, while others consider it at the document level, then aggregate similarities of documents in the corpus \cite{Otero2011}. The authors in \cite{Li2010} used a bilingual dictionary in their corpus level similarity measure. The dictionary is used to inspect the translation of each word of the source vocabulary in the target vocabulary. The objective is to improve the quality of the comparable corpus to extract a bilingual lexicon of good quality. The authors in \cite{Otero2011} reported that document level cross-lingual similarity measure can be useful to extract comparable corpora or to extract bilingual lexicon. The authors considered internal links of Wikipedia articles as a vocabulary. Internal links in Wikipedia articles are links to other Wikipedia articles. The authors considered these terms, which have internal links, as important terms in the documents. Thus, the source and the target documents are composed of internal links terms. The authors considered that two documents are comparable if they have most of these common (translation of each others) internal links.

The drawbacks of the dictionary based approach are the dependency on the bilingual dictionaries, which are not always available, and the necessity to use morphological analyzers for inflected languages. Moreover, word-to-word dictionary translations without considering the context can lead to many errors because of the multiplicity of senses (translation ambiguity), and because the text is considered only as a bag of independent words. 

In our work, we investigate a cross-lingual document similarity measured based on WordNet bilingual dictionary for English-Arabic documents. The similarity is measured in our work at document level. Further, two weighting schemes (binary and $\mathit{tfidf}$) are investigated.

\subsection{Cross-Lingual Information Retrieval (CL-IR) Methods}

Information Retrieval (IR) consists in finding documents in a large collection that are relevant to a given query composed of keywords \cite{IRbook}. Finding comparable documents is a very similar task to Information Retrieval (IR). To find comparable documents, the query is the whole document instead of keywords, and the task then is to find the most similar (relevant) document(s) to a given one.

The IR system usually computes a numerical score that quantifies how each document in the collection matches a query, and ranks the documents according to this value \cite{IRbook}. A document is ranked by estimating the cosine similarity with the query. The top ranked documents are then shown to the user. The IR system normally indexes text documents by representing them as vectors in the vector space.

Cross-Lingual Information Retrieval (CL-IR) is a special case of IR, where the language of the query is different from the language of the documents. In this case, the CL-IR system should unify the language of queries and documents. Therefore, the system should help users to understand the results regardless of the source language.

In CL-IR methods, either queries or documents can be translated using a Machine Translation (MT) system. Then, classical IR tools, which are based on Vector Space Model, can be used to identify comparable articles. The drawback of this approach is the dependency on the MT system, which affects the performance of the IR system. Moreover, the MT system needs to be developed first if it is not available for the desired language.

Researchers usually translate queries into the language of the indexed documents instead of translating the whole document collection \cite{Aljlayl2002}. This is because the computational cost of translating queries is far less than the cost of translating all indexed documents. Thus, the IR system indexes the target documents and the source queries are translated by the MT system. The authors in \cite{Aljlayl2002} reported that the approach is limited by the quality of the translation. In \cite{Ture2013}, the author addressed the problem of improving the performance of MT to have better retrieval results. His solution is to integrate IR and MT by tuning the MT system to improve the IR results.

In our work, we investigate different approaches for CL-IR. We investigate bilingual dictionary approach and machine translation approach for cross-lingual documents retrieval, but we investigate machine translation approach with Latent Semantic Indexing (LSI) method. We further compare the results of these approaches.

\subsection{Cross-Lingual Latent Semantic Indexing (CL-LSI) Methods} 

Document similarity can be estimated on term or semantic level \cite{Harispe2015}. Generally, semantic similarity is a metric that quantifies the likeness of documents based on the meaning or semantics of the contents. Semantically related terms are usually referred as \textit{concepts}. Semantic similarity can be measured based on a pre-defined ontology, which specifies the distance between concepts, or can be measured using statistical methods, which find correlation between terms and contexts in a text corpus. One of the methods in the latter category is Latent Semantic Indexing (LSI). 

LSI is designed to solve the problems of the ordinary IR system, which depends on lexical matching \cite{Dumais2007}. For instance, some irrelevant information can be retrieved because some literal words have different meanings. On the contrary, some relevant information can be missed because there are different ways to describe an object. The VSM in the ordinary IR system is sparse (most of elements are zeros), while LSI space is dense. In the sparse VSM, terms and documents are loosely coupled, while terms and documents in LSI space are coupled with weights. VSM is usually a high dimensional space for large documents collection, while LSI has the particularity to reduce the space, i.e., the number of dimension in LSI is lower than the number of unique terms in the document collection. In sparse and high dimensional space, cosine similarity can be noisy or inaccurate. 

Documents and words in LSI are projected into a reduced space using Singular Value Decomposition (SVD). Similar words and document are mapped closer to each others in the LSI space. LSI can capture  synonyms but it can not address polysemy.

The authors in \cite{Blei2003} extended the LSI to a probabilistic version called Latent Dirichlet Allocation (LDA). In LDA, documents are represented as random mixtures of latent topics, these topics are probabilistic distributions over words. Each document can be perceived as a mixture of different topics, and every topic is characterized by a distribution over words. LDA is useful for modeling topic that are mentioned in a corpus, while LSI is useful to map similar documents and words in a corpus into a reduced feature space (model concepts) \cite{Cui2011}.

In our work, we use LSI method for cross-lingual similarity measure because our aim is to map similar documents and words across languages closer to each others into a reduced feature space. In other words, we aim to model \textit{concepts} rather than \textit{topics}.

In Cross-Lingual Latent Semantic Indexing (CL-LSI) method, documents are represented as vectors like in CL-IR method, but these vectors are further transformed into another reduced vector space like in LSI. Then, one can compute the cosine distance between vectors in this new space to measure the similarity between them. LSI method has already been used for CL-IR in \cite{Littman1998}. In this approach, the source and the target documents are concatenated into one document. Therefore, LSI learns the links between source and target words. The CL-LSI method is described in detail in Section \ref{sec:cl_lsi_method}. The advantage of CL-LSI is that it does not need morphological analyzers or MT systems. Moreover, it overcomes the problem of vocabulary mismatch between queries and documents. \cite{Dumais2007} compared the performance of CL-IR and CL-LSI, and the authors showed that LSI outperforms the vector space model for IR.

Many works have been done on retrieval of document pairs written in various languages using CL-LSI. For example, \cite{Berry1995} worked on Greek-English documents and \cite{Littman1998} worked on French-English documents, Spanish-English \cite{Evans1998}, Portuguese-English \cite{Orengo2003}, Japanese-English \cite{Mori2001}, while \cite{Muhic2012} worked on several European languages.

In \cite{Littman1998}, the authors reported that CL-LSI performs well (98\% accuracy) for cross-lingual retrieval. They also reported that machine translation can be sufficient for cross-lingual retrieval using LSI. The authors also concluded that domain consistency between training and test texts is important for cross-lingual document retrieval using LSI. 

The authors in \cite{Muhic2012} applied CL-LSI method for seven European languages for document pair retrieval, and they reported that the CL-LSI method is language independent, but the performance is language dependent, i.e., it depends on the language pair.

The authors in \cite{Cimiano2009} conducted an empirical comparison between LSI, LDA, and Explicit Semantic Analysis (ESA) for CL-IR task. ESA method indexes the text using given external concepts or knowledge-base. The concepts are explicit in ESA, while the concepts are latent in LSI. In other words, the ESA's concepts are defined explicitly, while LSA's concepts are extracted implicitly (latent) from the corpus. The authors applied LSI, LDA, and ESA methods on two parallel corpora: the first corpus is collected from the ``Journal of European Community" (JRC-Acquis)\footnote{\url{http://langtech.jrc.it/JRC-Acquis.html}}, and the second one is collected from legislative documentations of the European Union (Multext)\footnote{\url{http://aune.lpl.univ-aix.fr/projects/MULTEXT/}}. The authors conducted experiments on three language pairs of these corpora; English-French, English-German, and French-German. JRC-Acquis and Multext corpora are split into training (60\%) and testing (40\%) parts. The task is to retrieve document pairs of the test parts of the parallel corpora. The LDA and LSI models are trained on the training parts of parallel corpora, while  Wikipedia articles is used as external knowledge-base for the ESA method, where each article is considered as a concept. The authors repeated the same experiment but they trained LSI and LDA on Wikipedia corpus instead of the training parts of JRC-Acquis and Multext parallel corpora. The authors experimentally determined the optimal dimensions for LSI/LDA (500), and  ESA (10,000). Their results indicate that performance of LDA/LSI is better when they are trained on the training parts rather than they are trained on Wikipedia. The results also showed that the performance depends on the method, the corpus, and the language pair. The general conclusions that can be drawn from their results are that ESA and LSI perform well and mostly have close performance, while the performance of LDA is poor. Finally, the authors claimed that ESA outperforms LSI/LDA when the latter are trained on Wikipedia instead of the training parts of JRC-Acquis and Multext parallel corpora, but this is expected because of the vocabulary mismatch problem between Wikipedia and JRC-Acquis and Multext corpora, and because Wikipedia is not fully parallel. In our work, we train LSI on EURONEWS corpus and use it to align different corpus (BBC-JSC), but  they all belong to the same domain (news articles).

The authors in \cite{Mccrae2013} combined ESA and LSI methods for cross-lingual document pairs retrieval. Their approach builds a set of explicitly defined topics and computes the latent similarity between these topics. The main idea is to map documents into a latent topic space, then into an explicit topic space. The authors claimed that their approach outperforms LSI, LDA, ESA methods. The authors chose the optimal number of dimensions for LDA experimentally, while they chose optimal number of dimensions for LSI according to the memory limit of their machine. Therefore, their claims that their approach outperforms other methods are questionable.

To summarize the works reviewed in this section, CL-LSI can achieve good results for cross-lingual document retrieval task \cite{Dumais2007}. Machine translation can be sufficient for cross-lingual document retrieval using LSI \cite{Littman1998,Fortuna2005}, but the benefit of CL-LSI is that it does not need machine translation and it has a competitive performance. The CL-LSI method is language independent, but the performance is language dependent, i.e., it depends on the language pair \cite{Muhic2012}. To achieve better results when using CL-LSI, it is recommended that the training and test documents are from close domains \cite{Littman1998,Fortuna2005,Cimiano2009}. ESI and LSI perform well for document pair retrieval task, and they have close performance, but LDA has poor performance for the same task \cite{Cimiano2009}.


In this work we investigate cross-lingual document pairs retrieval and alignment using CL-LSI but for Arabic-English languages. We also investigate the performance of retrieving documents using machine translation approach, and compare it to CL-LSI. In this paper, we investigate training CL-LSI using two types of corpora (parallel and comparable). In our work we use CL-LSI for two tasks: (1) document pair retrieval, and (2) aligning comparable documents. In the first task, the corpus is already aligned and the objective is to retrieve the target document by providing the source document as a query. In this task we apply the similarity measures on several corpora and we compare their performance. In the second task, the corpus is not aligned, and the objective is to automatically align comparable source and target documents which are related to the same context.

\section{Collected and Used corpora}
\label{sec:corpora}

This section presents the English-Arabic parallel corpora that we use in this work, and the English-Arabic-French comparable corpora that we collect. We need the parallel corpora in this work for several reasons: (a) to compare the application of the proposed methods on comparable as well as on parallel corpora, (b) to study the degree of similarity of comparable texts as compared to the degree of similarity of parallel texts. 

English-French-Arabic comparable corpora are not available. Therefore, collecting such resources is one of the contributions in this research. In addition, in this work we need such dataset to study comparable texts, and to develop and test our proposed methods for aligning and retrieving these documents. Moreover, such resources can be useful for several applications such as cross-lingual text mining, bilingual lexicon extraction, cross-lingual information retrieval and machine translation.

It should be noted that we collected the comparable corpora in Arabic, English and French languages, but we focus on English-Arabic language pair for alignment task. Before describing the used and collected corpora, we briefly introduce some characteristics of the Arabic language in the next section.

\subsection{Arabic Language}
\label{sec:arabic}

\transtrue 
Arabic language is used by about 422M people in the Middle East, North Africa and the Horn of Africa \cite{Arabic}.  Arabic words are derived from roots, which can be composed of three, four or five letters. Triliteral root is the most common one. About 80\% of Arabic roots are triliteral \cite{Khoja1999stemming,Sawalha2008}. Words can be derived from the root by adding prefixes, infixes or suffixes.

Arabic is a highly inflected language \cite{Habash2010}. Table \ref{tab:roots} presents some examples of inflected terms of the Arabic language. The table shows several different English words, that are related to the same root in Arabic. Therefore, for an English-Arabic NLP task, applying rooting on Arabic words may lead to lose the meaning of Arabic words against the corresponding English words.

\begin{table}[!htb]
\caption{Examples of some inflected terms in Arabic language}
\label{tab:roots}
\centering
\begin{tabular}{|c|c|l|c|}
\hline
Arabic word & Meaning & Description & Root \\
\hline

\RL{||كاتب} &  author & name of the subject & \RL{||كتب}  \\ 
\RL{||يكتب} & he writes & the verb & \RL{||كتب}  \\
\RL{||كتاب} &  book & the outcome of the verb & \RL{||كتب}  \\
\RL{||مكتبة} &  library & where the verb takes place & \RL{||كتب}   \\
\RL{||مكتب} &  office & the place of the verb (to write) & \RL{||كتب}  \\

\hline

\RL{||يطير} & he flies & the verb & \RL{||طير}  \\
\RL{||طائر} &  bird &  name of the subject & \RL{||طير}  \\
\RL{||طيار} &  pilot &  name of the subject & \RL{||طير}  \\
\RL{||طائرة} &  airplane &  name of the subject & \RL{||طير}   \\

\hline

\end{tabular}
\end{table}

Unlike English terms that are isolated, certain Arabic terms can be agglutinated (words or terms are combined) \cite{Habash2010}. For instance, the Arabic item \RL {||وسيعطيك} corresponds to ``and he will give you" in English.  In Arabic, usually the definite article \RL {||الـ} ``the" and pronouns are connected to the words. Arabic words have different forms depending on gender (masculine and feminine). For example, the English word ``travelers" corresponds to  \RL {||مسافرون} in masculine form, and \RL {||مسافرات} in feminine form. Word forms in Arabic may change according to its grammatical case. For instance, \RL {||مسافرون} is in nominative form, while its accusative/genitive form is \RL {||مسافرين}. Besides the singular and plural forms, Arabic words have the dual form. Singular form refers to one person or thing, dual form refers to two persons or things, and plural form refers to three or more persons or things. Verb conjugation in Arabic is derived according to person, number, gender, and tense. See \cite{Saad2015phd} for more details.

Common methods to analyze words in Arabic language is rooting \cite{Khoja1999stemming,Taghva2005} and light stemming \cite{Larkey2007Light}. Rooting removes the word's prefix, suffix and infix, then converts it into the root form, while light stemming just removes the word's prefix and suffix. Table \ref{tab:rooting_vs_light_stemming} shows some examples, where words are analyzed using rooting and light stemming methods.

\begin{table}[!htb]
\caption{Methods of morphology analysis for some Arabic words}
\label{tab:rooting_vs_light_stemming}
\centering
\begin{tabular}{|c|c|c|c|c|c|}
\hline 
Word & Meaning & Prefix & Infix & Suffix & Light Stem \\ 
\hline 
\multicolumn{6}{|l|}{Words inflected  from the root \RL{||كتب}  (to write)} \\
\hline 
\RL{||المكتبة}  & the library & \RL{||الـ} & \RL{||ـمـ} & \RL{||ـة}  & \RL{||مكتب} \\ 
\RL{||الكاتب}  & the author & \RL{||الـ} & \RL{||ـا} & - &  \RL{||كاتب} \\ 
\RL{||الكتاب}  & the book & \RL{||الـ} & \RL{||ـا} & - &  \RL{||كتاب} \\ 
\RL{||يكتب}  & he writes & \RL{||يـ} & - & - &  \RL{||كتب} \\ 
\hline
\multicolumn{6}{|l|}{Words inflected  from the root \RL{||سفر}  (to travel)} \\
\hline 
\RL{||المسافرون}  & the travelers & \RL{||المـ} & \RL{||ـا} & \RL{||ون} & \RL{||مسافر} \\ 
\RL{||المسافرين}  & the travelers & \RL{||المـ} & \RL{||ـا} & \RL{||ين} & \RL{||مسافر} \\ 
\RL{||سيسافر}  & he will travel & \RL{||سيـ} & \RL{||ـا} & - & \RL{||سافر} \\ 
\RL{||سافرت}  & she traveled & - & \RL{||ـا} & \RL{||ت} & \RL{||سافر} \\ 
\hline 
\end{tabular} 
\end{table}

Light stemming have better performance than rooting for several Arabic NLP tasks such as text classification \cite{Saad2010msc}, text clustering \cite{Ghanem2014}, information retrieval \cite{Larkey2007Light}, and measuring texts similarity \cite{Froud2012}. For English-Arabic NLP tasks, applying rooting on Arabic words may lead to lose the meaning of Arabic words against the corresponding English words \cite{Saad2013cilc}.  

To recapitulate, Arabic language has very different characteristics from English language. Several consideration should be taken into account when doing Arabic or Arabic-English NLP tasks. This makes the task more challenging.

\subsection{Comparable Corpora}
\label{sec:comparable_corpus}

\transfalse
This section describes the comparable corpora that we collect from two sources: EURONEWS website\footnote{\url{www.euronews.com}}, and Wikipedia encyclopedia\footnote{\url{www.wikipedia.org}}. We align the collected texts at the document level. That means, for EURONEWS corpus, that aligned articles are related to the same news story, and for Wikipedia corpus, that aligned articles are related to the same context. For example, the English Wikipedia article ``Olive oil" is aligned to the French article ``Huile d'olive", and to the Arabic article ``\RL{زيت زيتون}". In the next two sections, we described our collected comparable corpora. 

\subsubsection{Wikipedia Comparable Corpus}

Wikipedia is an open source encyclopedia written by contributors in several languages. Anyone can edit and write Wikipedia articles. Therefore, articles are usually written by different authors. Some Wikipedia articles of some languages are translations of the corresponding English versions, while other articles are written independently. Wikipedia provides a free copy of all available contents of the Encyclopedia (articles, revisions, discussion of contributors). These copies are called dumps\footnote{\url{dumps.wikimedia.org}}. Because Wikipedia contents change with time, the dumps are provided regularly every month. Wikipedia dumps can be downloaded in XML format. Our Wikipedia corpus is extracted by parsing Wikipedia dumps of December 2011, which are composed of 4M English, 1M French, and 200K Arabic articles.

Arabic, French, and English comparable articles are extracted based on inter-language links. In a given Wikipedia article written in a specific language, ``inter-language links" refer to the corresponding articles in other languages. The form of these links is $[[languagecode:Title]]$. For example, for Wikipedia article which is related to the biography of ``Lawrence", the link for the English article is [[en:T. E. Lawrence]], and the link for the French article is [[fr:Thomas Edward Lawrence]].

Using inter-language links for a given Wikipedia articles, we can select the titles of Wikipedia documents in other languages and extract them and link (align) them together. Thus, the extracted articles are aligned at article level. That means the three comparable articles are related to the same topic (context). We denote the extracted corpus as Arabic-French-English Wikipedia Corpus (AFEWC). The following steps describe our approach to extract and align comparable articles from Wikipedia dumps. These steps are applied for each English article in Wikipedia dump files.

\begin{enumerate}
\item If the English article contains Arabic and French inter-language links, then extract the French and Arabic titles from their inter-language links.
\item Search for these titles in the Wikipedia dump XML file to extract their corresponding articles. 
\item Extract the plain-text of the three comparable articles from wiki-markup. 
\item Write comparable articles in plain-texts and xml formats.
\end{enumerate}

The extracted information includes article's title and wiki markup. From wiki markup, we extract the article's summary (abstract), categories, and the plain text. Examples of generic categories are sport, economics, religion, etc. Examples of specific categories are `Nobel Peace Prize laureates', `cooking oils', etc. All the aligned articles are structured in XML files. We also keep the wiki-markup for the aligned articles because it can be useful to extract additional information later such as info boxes, image captions, etc.

Wikipedia December 2011 dumps contain about 4M English articles, 1M French articles, and 200K Arabic articles. In total, we extracted and aligned about 40K comparable articles. Corpus information is presented in Table \ref{tab:wikipedia_corpus}, where $\vert D \vert$ is the number of articles, $\vert S \vert$ is the number of sentences, $\vert W \vert$ is the number of words, $\vert V \vert$ is the vocabulary size, $\bar{S}$ is the average number of sentences per article, and $\bar{W}$ is average words per article. It can be noted from Table \ref{tab:wikipedia_corpus} that the number of sentences of Arabic articles is less than the number of sentences of English and French articles.

\begin{table}[!htb]
\caption{Wikipedia comparable corpus (AFEWC) characteristics}
\label{tab:wikipedia_corpus}
\centering
\begin{tabular}{|c|c|c|c|}
\hline 
 & English & French & Arabic \\
\hline  
$\vert D \vert$  & 40K & 40K & 40K \\ 
$\vert S \vert$  & 4.8M & 4.7M & 1.2M \\ 
$\vert W \vert$ & 91.3M & 57.8M  & 22M \\ 
$\vert V \vert$ & 2.8M & 1.9M & 1.5M \\
{\small $\bar{S}$}  & 119 & 69 & 30 \\
{\small $\bar{W}$ }  & 2.2K & 1.4K & 548 \\
\hline 
\end{tabular}
\end{table}

\subsubsection{EURONEWS Comparable Corpus}

EURONEWS is a multilingual news TV channel, which aims to cover world news from a pan-European perspective\footnote{\url{www.euronews.com}}. News stories are also posted on the website. EURONEWS is available now in many European languages as well as in Arabic. English and French news services started in 1993, while Arabic started in 2008. 

EURONEWS corpus is extracted by parsing the html files of articles collected from EURONEWS website. Each English document has a hyperlink to the corresponding Arabic and French articles. We align comparable articles using these hyperlinks. Then, html tags are stripped for the three comparable articles, and stored in plain text files. Category information is also included in the plain text files. EURONEWS categories are: cinema, corporate, economy, Europe, hi-tech, interview, markets, science, and world. The corpus contains about 34K comparable articles as shown in Table \ref{tab:euronews_corpus}. The average number of sentences is almost the same in English, French and Arabic documents.

\begin{table}[!htb]
\caption{EURONEWS comparable corpus characteristics}
\label{tab:euronews_corpus}
\centering
\begin{tabular}{|c|c|c|c|}
\hline 
 & English & French &  Arabic \\
\hline 
$\vert D \vert$  & 34K & 34K & 34K \\
$\vert S \vert$  & 744K & 746K & 622K \\
$\vert W \vert$ &  6.8M & 6.9M  & 5.5M \\
$\vert V \vert$ &  232K & 256K & 373K \\
{\small $\bar{S}$}  & 21 & 21 & 17 \\
{\small $\bar{W}$ }  & 198 & 200 & 161 \\
\hline 
\end{tabular}
\end{table}


\subsection{Parallel Corpora}
\label{sec:parallel_corpus}

In this research we work on several corpora in order to measure the robustness of the studied methods. In this section we describe the parallel corpora that we use in this work.

We need parallel corpora in our work because it will be considered as a kind of baseline reference. In fact, testing the comparability measures on parallel corpora must give better results than those on comparable corpora. The parallel corpora come from several different domains, and they are ideal to test each method on different genres of texts.

\begin{table}[!htb]
\caption{Parallel Corpora characteristics}
\label{tab:corpus_parallel}
\centering
\begin{tabular}{|c|c|c|c|c|c|}
\hline 
\multirow{2}{*}{\bf Corpus} & \multirow{2}{*}{$\vert S \vert$} & \multicolumn{2}{|c|}{$\vert W \vert$} & \multicolumn{2}{|c|}{$\vert V \vert$}  \\ 
  &   &  English & Arabic &  English & Arabic \\
\hline 
\multicolumn{6}{|l|}{\bf Newspapers}    \\    
\hline          
AFP &   4K & 140K & 114K & 17K & 25K  \\ 
ANN &   10K & 387K & 288K & 39K & 63K \\ 
ASB &   4K & 187K & 139K & 21K & 34K \\ 
Medar &  13K & 398K & 382K & 43K & 71K \\ 
NIST &   2K & 85K & 64K & 15K & 22K \\ 
\hline
\multicolumn{6}{|l|}{\bf United Nations Resolutions}    \\              
\hline
UN &    61K & 2.8M & 2.4M & 42K & 77K \\ 
\hline
\multicolumn{6}{|l|}{\bf Talks}    \\   
\hline           
TED &   88K & 1.9M & 1.6M & 88K & 182K \\ 
\hline
\multicolumn{6}{|l|}{\bf Movie Subtitles}    \\ 
\hline             
OST &    2M & 31M & 22.4M & 504K & 1.3M \\ 
\hline
\hline
{\bf Total}  & 2.2M & 36.8M & 27.4M & 769K & 1.8M \\ 
\hline
\end{tabular} 
\end{table}

Table~\ref{tab:corpus_parallel} shows the characteristics of the parallel corpora that we use in this work. $\vert S \vert$ is the number of sentences, $\vert W \vert$ is the number of words, and $\vert V \vert$ is the vocabulary size. The table also shows the domain of each corpus. The parallel corpora are:
AFP\footnote{\url{www.afp.com}}, ANN\footnote{\url{www.annahar.com}}, ASB\footnote{\url{www.assabah.com.tn}} \cite{Dalal2009}, Medar\footnote{\url{www.medar.info}}, NIST  \cite{NIST}, UN \cite{Rafalovitch2009}, TED\footnote{\url{www.ted.com}} \cite{Cettolo2012} and OST\footnote{\url{www.opensubtitles.org}} \cite{Tiedemann2012}. The corpora are collected from different sources and present different genres of text. It is remarkable that AFP, ANN, ASB, Medar, NIST and UN corpora are generated by professional translators, while TED and OST are generated by volunteer translators.

As can be noted from Table~\ref{tab:corpus_parallel}, in all parallel corpora, English texts have more words than Arabic ones. The reason is that certain Arabic terms can be agglutinated, while English terms are isolated, as described in Section \ref{sec:arabic}. In contrast, the vocabulary of Arabic texts is larger than the vocabulary of the English one. This is because Arabic is a highly inflected language, as described in Section \ref{sec:arabic}.

\section{Cross-lingual Similarity Measures}
\label{sec:similarity_measure}

In this section, we present two cross-lingual similarity measures: the first one uses a bilingual dictionary and the second one is developed using Latent Semantic Indexing (LSI) method. We use these measures for document pairs retrieval, i.e., to retrieve a target document by using the source document as a query. In our work, we focus on English-Arabic language pair. We evaluate these methods on several parallel corpora, then we compare and discuss the performance of these methods. Finally, we use the best method to align further comparable documents collected from sources different of Wikipedia or EURONEWS. Namely, we align news documents collected from the British broadcasting corporation (BBC) and ALJAZEERA news agencies.

A document can be more or less similar to other documents, and we need a measure that can identify the degree of similarity of these documents. A similarity measure is a real-valued function that quantifies the likeness of the meaning or the contents of two documents. The function estimates the distance between two units of text (terms, sentences, paragraphs, documents, or concepts) through numerical representations of the text documents. The value of these measures range from 1 (exactly similar) to 0 (not similar). In the following sections, we present our measures, our experiment setup, then we discuss and compare the results.

\subsection{Cross-lingual Similarity Using Bilingual Dictionary}
\label{sec:dictionary_method}

In dictionary-based methods, the source and the target documents are comparable if most of words in source are translations of words in target \cite{Li2010}. Our dictionary based method uses multi-WordNet bilingual dictionary \cite{Bond2012} to match source and target words of the comparable documents. This method requires the source and target texts to be represented as vectors of matched words. For inflected languages, bilingual dictionaries usually do not cover all word variations, so morphological analysis is applied on words to improve the dictionary coverage. 

A document is represented by a vector made up of one feature (or weight) per word. Word weights can be either binary (1 or 0 to indicate the presence or absence of the translation in the target document) or numerical represented by the term frequency-inverse document frequency ($\mathit{tfidf}$) of words in the document. 

In order to measure the similarity between two documents, we compare their vectors. For binary weighting scheme we propose a binary measure, and for $\mathit{tfidf}$ weighting scheme we propose a cosine measure. For a given source document $d_s$ and target document $d_t$, the binary measure counts the words in $d_s$ which have translations in $d_t$ and then normalizes these counts by the vector size, while the cosine measure computes the cosine similarity between source and target vectors which are represented by the $\mathit{tfidf}$ of the matching words of $d_s$ and $d_t$. The binary measure  uses the function $trans(w_s,d_t)$, which returns 1 if a translation of the source word $w_{s}$ is found in the target document $d_t$, and 0 otherwise. The similarity using the binary measure can be computed as follows:

\begin{equation}
\label{eq:binCM}
bin(d_s,d_t)=\frac{\underset{w_s \in d_s \cap V_s}{\sum} trans(w_s,d_t)}{|d_s \cap V_s|}
\end{equation}

where $V_s$ is the source vocabulary of the bilingual dictionary, $d_s$  and $d_t$ are the source and target documents considered as bags of words. Because $bin(d_s,d_t)$ is not symmetric, the actual value used for measuring the comparability between $d_s$ and $d_t$ is as follows:

\begin{equation}
\label{eq:binCM_sym}
\frac{bin(d_s,d_t)+bin(d_t,d_s)}{2}
\end{equation}

Cosine similarity is a measure of similarity, between two vectors in a vector space, which measures the cosine of the angle between the two vectors. Source and target texts can be represented as vectors where the value of each dimension corresponds to weights/features (e.g. $\mathit{tfidf}$) associated to the matched words in the documents. This representation is generally referred to as a Vector Space Model (VSM). Given two vectors $d_s$ and $d_t$ of $n$ attributes representing  the source and target documents, the cosine similarity $cosine(d_s,d_t)$ between these documents is computed as follows:

\begin{equation}
\label{eq:cosine}
cosine(d_s,d_t)=\frac{d_s \cdot d_t}{\Vert d_s \Vert \times  \Vert d_t \Vert}
=
\frac{\overset{n}{\underset{i=1}{\sum}}d_{s_i}\times d_{t_i}}
{
\sqrt{\overset{n}{\underset{i=1}{\sum}}(d_{s_i})^{2}}
\times 
\sqrt{\overset{n}{\underset{i=1}{\sum}}(d_{t_i})^{2}}
}
\end{equation}

To represent cross-lingual documents in the VSM, we build the source and target vectors as follows: using a bilingual dictionary, for each translation $w_s \leftrightarrow w_t$ in this dictionary, define one attribute of the vectors. For the source vector this attribute is equal to the $\mathit{tfidf}$ of $w_s$ (0 if $w_s$ is not in the source document), and for the target vector this attribute is equal to the $\mathit{tfidf}$ of $w_t$ (0 if $w_t$ is not in the target document).

We use the Open Multilingual WordNet (OMWN) bilingual dictionary \cite{Bond2012} in our work to match the source and the target words. OMWN is available in many languages including Arabic and English. OMWN has 148K English words and 14K Arabic words. Synonym words are grouped into sets called synsets. These synsets help to identify possible translations from source to target. To match words in the source and the target texts, each word is looked up in the bilingual dictionary. Before that, morphological analysis is applied on words to increase the coverage of dictionary between source and target texts. Also stop words and punctuation are removed from all the texts before matching words. 

There are many word reduction techniques for English and Arabic languages. For English, stemming and lemmatization are widely used in the community. Stemming \cite{Porter2001} prunes a word into a stem, which is a part of the word, and may not be in the dictionary, while lemmatization \cite{Miller1998} retrieves the dictionary form (lemma) of an inflected word. As for Arabic, rooting \cite{Khoja1999stemming,Taghva2005} or light stemming \cite{Larkey2007Light} are widely used techniques. Rooting removes the word's prefix, suffix and infix, then converts it to the root form, while light stemming just removes the word's prefix and suffix. As discussed in Section \ref{sec:arabic}, Arabic is a highly inflected language. Thus, applying rooting leads to lose the meaning of Arabic words against the corresponding English words. For more details about rooting and light stemming, see \cite{Saad2015phd}.

In order to increase English-Arabic word matching using the bilingual dictionary, we have developed a new reduction technique for Arabic words, which combines rooting and light stemming techniques \cite{Saad2013cilc}. We name this technique as \textit{morphAr}. The idea is to try to reduce Arabic words by applying light stemming first, and then applying rooting. If the stem is found in the dictionary, then its translations are returned, otherwise the translations of the root are returned.

We have two reduction techniques for English (stemming and lemmatization) and three techniques for Arabic (light stemming, rooting and \textit{morphAr}). To determine the best combination of these techniques, we conducted an experiment using each technique separately, also inspecting the percentage of words that are Out Of Vocabulary (OOV). This experiment is applied on AFP, ANN, ASB, TED, UN parallel corpora, which are described in Section \ref{sec:parallel_corpus}. The OOV rate is computed as follows:

\begin{equation}
\label{eq:oov}
 \frac{1}{2} \times \big( \frac{ \vert w_{s}^{oov} \vert }{\vert d_s \vert}  + \frac{ \vert w_{t}^{oov} \vert }{\vert d_t \vert} \big)
\end{equation}

where $d_s$ is the source document, $d_t$ is the target document, $\vert d \vert$ is the word count in the document and $\vert w_{}^{oov} \vert$ is the count of the words that are OOV (not found in the dictionary).

\begin{figure}[!htb]
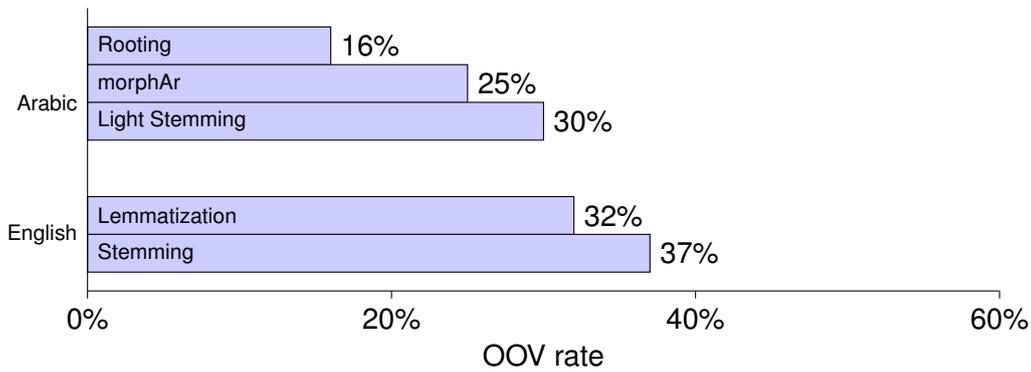

\centering  
\begin{bchart}[max=60,step=20,unit=\%,width=12cm]

\bcbar[text={\scriptsize Rooting}]{16}
\bcbar[text={\scriptsize morphAr}]{25}
\bclabel{{\scriptsize Arabic}}
\bcbar[text={\scriptsize Light Stemming}]{30}

\bigskip

\bcbar[text={\scriptsize Lemmatization}]{32}
\bclabel{{\scriptsize English}}
\bcbar[text={\scriptsize Stemming}]{37}
\bcxlabel{OOV rate}
\end{bchart}
\caption{OOV rate using different word reduction techniques for Arabic and English parallel corpus}
\label{fig:morph-standalone}
\end{figure}

Figure \ref{fig:morph-standalone} shows the OOV rate using different word reduction techniques for Arabic and English parallel corpus. The figure presents the OOV rate for each word reduction technique separately. If we consider word reduction techniques for each language separately, then rooting for Arabic and lemmatization for English have the lowest OOV rate as shown in Figure \ref{fig:morph-standalone}. But we do not aim to just reduce OOV independently for each language. Instead, we aim to increase matching rate of source and target word translations by finding the appropriate translation for these words using the bilingual dictionary. Let $\vert w_s \leftrightarrow w_t \vert$ be a matching between a source word $w_s$ and its translation $w_t$, then word matching rate is the count of source and target words that are translation of each others in the source and the target documents ($\vert w_s \leftrightarrow w_t \vert$), normalized by source and target document sizes $\vert d_s \vert$ and $\vert d_t \vert$ respectively, and it is computed as follows:

\begin{equation}
\label{eq:matching_rate}
\frac{ \vert w_s \leftrightarrow w_t \vert }{\vert d_s \vert + \vert d_t \vert}
\end{equation}

\begin{figure}[!htb]
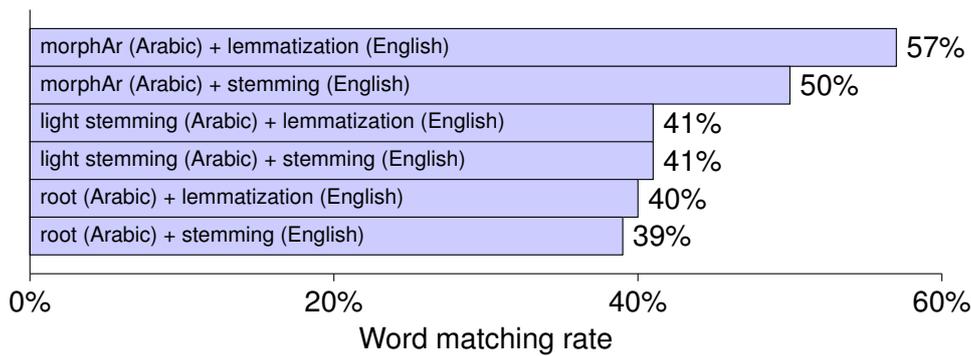

\centering  
\begin{bchart}[max=60,step=20,unit=\%,width=12cm]
\bcbar[text={\scriptsize morphAr (Arabic) + lemmatization (English)}]{57}
\bcbar[text={\scriptsize morphAr (Arabic) + stemming (English)}]{50}
\bcbar[text={\scriptsize light stemming (Arabic) + lemmatization (English)}]{41}
\bcbar[text={\scriptsize light stemming (Arabic) + stemming (English)}]{41}
\bcxlabel{Word matching rate}
\bcbar[text={\scriptsize root (Arabic) + lemmatization (English)}]{40}
\bcbar[text={\scriptsize root (Arabic) + stemming (English)}]{39}
\end{bchart}
\caption{Word matching rate of combined Arabic-English word reduction techniques using the bilingual dictionary}
\label{fig:morph}
\end{figure}

The word matching rates for different combinations of the word reduction techniques in both Arabic and English are presented in Figure \ref{fig:morph}. It can be noted that \textit{morphAr} for Arabic and lemmatization for English lead together to the best coverage (best matching rate). Therefore, we use this combination of techniques in our experiments. As shown in Figure \ref{fig:morph}, rooting for Arabic with other English word reduction techniques has the lowest matching rate. Recall from Section \ref{sec:corpora}, using rooting in Arabic language leads to lose the corresponding meaning in the English language.

In the next section, we present LSI based measure, then we present experimental results of applying the dictionary based and the LSI based measures on parallel corpora.

\subsection{Cross-lingual Similarity Using CL-LSI}
\label{sec:cl_lsi_method}

In this section we present a cross-lingual similarity measure based on the Cross-Lingual Latent Semantic Indexing (CL-LSI). 

In our work, we use the same approach as \cite{Littman1998}, but we apply it on Arabic-English documents. Moreover, \cite{Littman1998} used parallel corpus to train the CL-LSI, whereas we use both parallel and comparable corpora for training. 

We describe below formally how we deal with parallel/comparable corpus, how we extract matrices from these corpus and how we use the Latent Semantic Indexing approach in order to define a measure of similarity between documents.

We have a set of couples of documents $(a_j,e_j)$. $a_j$ is an Arabic document, $e_j$ is an English document. $a_j$ and $e_j$ are comparable or parallel (in this case, a document is actually a sentence). $A$ is the Arabic corpus, $E$ is the English corpus. $V_A$ is the vocabulary of the whole set of Arabic documents, and $V_E$ is the vocabulary of the whole set of English documents. There are $d$ couples of documents.

We describe this corpus by two matrices:

\begin{itemize}
\item $\mathcal{A}$ is the Arabic matrix, it is composed of $|V_A|$ lines and $d$ columns. Each line corresponds to a word $w_i$ of $V_A$. Each column corresponds to a document $a_j$. $\mathcal{A}_{ij}$ contains a value representing the presence of the word $w_i$ in the document $a_j$ (see Equation \ref{eq:A_mono_lingual}). 

\begin{equation}
\label{eq:A_mono_lingual}
\mathcal{A} = 
\bordermatrix{  
 & a_1 & a_2 & a_3 & \ldots & a_d \cr
w_1 & \mathcal{A}_{11} &  \mathcal{A}_{12}  & \mathcal{A}_{13}  & \ldots & \mathcal{A}_{1d} \cr
w_2 & \mathcal{A}_{21}  &  \mathcal{A}_{22} & \mathcal{A}_{23}  & \ldots & \mathcal{A}_{2d} \cr
w_3 & \mathcal{A}_{31}  &  \mathcal{A}_{32} & \mathcal{A}_{33}  & \ldots & \mathcal{A}_{3d} \cr
\vdots & \vdots & \vdots & \vdots & \ddots & \vdots \cr
w_{|V_A|} & \mathcal{A}_{|V_A|1}  &  \mathcal{A}_{|V_A|2} & \mathcal{A}_{|V_A|3}   & \ldots & \mathcal{A}_{|V_A|d}}_{d \times |V_A|}
\end{equation}

\item $\mathcal{C}$ (for Cross Lingual) is the English/Arabic matrix, itis composed  of $|V_E|+|V_A|$ lines and $d$ columns. Each line corresponds to a word $w_i$ of $V_E$ or $V_A$. Each column corresponds to a couple of documents  $(a_j,e_j)$. $\mathcal{C}_{ij}$ contains a value representing the presence of the word $w_i$ (english or arabic) in the couple of documents $(a_j,e_j)$ (in this case $(a_j,e_j)$ is considered as a simple document made up of the concatenation of $a_j$ and $e_j$) (see Equation \ref{eq:CL_cross_lingual}).

\begin{equation}
\label{eq:CL_cross_lingual}
\mathcal{C} = 
\bordermatrix{  
 & (a_1,e_1) & (a_2,e_2) & (a_3,e_3) & \ldots & (a_d,e_d) \cr
w_1 & \mathcal{C}_{11} &  \mathcal{C}_{12}  & \mathcal{C}_{13}  & \ldots & \mathcal{C}_{1d} \cr
w_2 & \mathcal{C}_{21}  &  \mathcal{C}_{22} & \mathcal{C}_{23}  & \ldots & \mathcal{C}_{2d} \cr
w_3 & \mathcal{C}_{31}  &  \mathcal{C}_{32} & \mathcal{C}_{33}  & \ldots & \mathcal{C}_{3d} \cr
\vdots & \vdots & \vdots & \vdots & \ddots & \vdots \cr
w_{|V_A|+|V_E|} & \mathcal{C}_{(|V_A|+|V_E|)1}  &  \mathcal{C}_{(|V_A|+|V_E|)2} & \mathcal{C}_{(|V_A|+|V_E|)3}   & \ldots & \mathcal{C}_{(|V_A|+|V_E|)d}}_{d \times (|V_A|+|V_E|)}
\end{equation}

\end{itemize}

The values in matrices are the $\mathit{tfidf}$ values of words into documents.


\vspace{1cm}

From each matrix, we build a LSI matrix. For that, we follow the classical method based on Singular Value Decomposition (SVD) \cite{Deerwester1990}. For example, we apply the SVD to $\mathcal{A}$ in order to obtain $U_\mathcal{A}S_\mathcal{A}^{}V_\mathcal{A}^t$ ($t$ is the transpose operator). $U_\mathcal{A}^{}$ is the term matrix composed of $V_A$ lines and $k$ columns, where $k$ is the reduced dimension in the LSI space.  \cite{Landauer1998,Dumais2007} reported that the optimal value of $k$ to perform SVD is between 100 and 500. Thus, one can determine the optimal value of $k$ between 100 and 500 experimentally. Each column vector in $U_\mathcal{A}^{}$ maps the terms of $V_A$ into a single concept of semantically related terms that are grouped with similar values. The diagonal matrix $S_\mathcal{A}^{}$ is composed of $k$ lines and $k$ columns of singular values. The document matrix $V_\mathcal{A}^{}$ is a $d \times k$ matrix.

Then, in the case of $\mathcal{A}$, an Arabic document $a$ is described by a vector $v_{a,\mathcal{A}}$ of $V_A$ parameters. It is possible to project this vector into the LSI space by the following formula:

\begin{equation}
\label{lsi_arabic}
v'_{a,\mathcal{A}} = v_{a,\mathcal{A}}^tU_\mathcal{A}^{} S_\mathcal{A}^{-1}
\end{equation}

$v'_{a,\mathcal{A}}$ is a vector of $k$ parameters. This vector describes the document $a$ in the LSI space.

Then, we define the similarity between two documents $a_1$ and $a_2$ by the cosine distance between $v'_{a_1,\mathcal{A}}$ and $v'_{a_2,\mathcal{A}}$.

\vspace{1cm}

But, actually, we want to measure the similarity between an Arabic document and an English document. For that, we use two strategies.

\paragraph{The monolingual strategy (AR-LSI):} the objective is to compare the documents $a$ (Arabic) and $e$ (English). For that, we follow the following steps:

\begin{enumerate}
\item we translate $e$ into Arabic, obtaining $e_a$
\item we describe $a$ and $e_a$ by vectors of the $\mathcal{A}$ space, obtaining $v_{e_a,\mathcal{A}}$ and $v_{a,\mathcal{A}}$
\item we project $v_{e_a,\mathcal{A}}$ and $v_{a,\mathcal{A}}$ into the Arabic LSI space by using  equation (\ref{lsi_arabic}). We obtain $v'_{e_a,\mathcal{A}}$ and $v'_{a,\mathcal{A}}$
\item we compute the cosine distance between $v'_{e_a,\mathcal{A}}$ and $v'_{a,\mathcal{A}}$.
\end{enumerate}

\paragraph{The cross-lingual strategy (CL-LSI):} the objective is to compare the document $a$ (Arabic) and $e$ (English).

For that, we build the LSI space corresponding to the Cross-Lingual matrix $\mathcal{C}$:

\begin{equation}
\label{cl}
\mathcal{C} = U_\mathcal{C}^{}S_\mathcal{C}^{}V_\mathcal{C}^t
\end{equation}

and we use the corresponding formula to project a $\delta$ document vector into the Cross-Lingual LSI space:

\begin{equation}
\label{lsi_cl}
v'_{\delta,\mathcal{C}} = v_{\delta,\mathcal{C}}^tU_\mathcal{C}^{} S_\mathcal{C}^{-1}
\end{equation}

Then, we follow the following steps:

\begin{enumerate}
\item we describe $e$ in terms of $\mathcal{C}$ obtaining $v_{e,\mathcal{C}}$. A parameter of $v_{e,\mathcal{C}}$ corresponding to an English word is the $\mathit{tfidf}$ value of this word into the English document; a parameter of $v_{e,\mathcal{C}}$ corresponding to an Arabic word is fixed to 0. 
\item we describe $a$ in terms of $\mathcal{C}$ obtaining $v_{a,\mathcal{C}}$. The parameter of $v_{a,\mathcal{C}}$ corresponding to an English word is fixed to 0; the parameter of $v_{a,\mathcal{C}}$ corresponding to an Arabic word is the $\mathit{tfidf}$ value of this word into the Arabic document; 
\item we project $v_{e,\mathcal{C}}$ and $v_{a,\mathcal{C}}$ into the Cross-Lingual LSI space by using formula (\ref{lsi_cl}). We obtain $v'_{e,\mathcal{C}}$ and $v'_{a,\mathcal{C}}$
\item we compute the cosine distance between $v'_{e,\mathcal{C}}$ and $v'_{a,\mathcal{C}}$.
\end{enumerate}

The advantage of cross-lingual LSI method is that it does not need bilingual dictionaries, morphological analyzers or machine translation systems. Moreover, this method overcomes the problem of vocabulary mismatch between queries and documents. This method allows to achieve our objective to retrieve comparable articles. We describe in the next section how we use the similarity measures in order to retrieve documents.


\subsubsection{Experiment Procedure}
\label{sec:experimental_procedure}

In order to evaluate the LSI-based similarity measures, we split the corpora into a training part (90\%) and a test part (10\%). We build the $\mathcal{A}$ (AR-LSI) and $\mathcal{C}$ (CL-LSI) matrices using the training part.

As mentioned earlier, the optimal value of $k$ (the LSI space size) for AR-LSI and CL-LSI can be chosen experimentally. To choose this value, we follow the experience of \cite{Landauer1998,Dumais2007}, who reported that the optimal value of $k$, to perform Singular Value Decomposition (SVD), is between 100 and 500. We conducted several experiments in order to determine the best rank for AR-LSI and CL-LSI, and we found that 300 optimizes the similarity for the parallel corpus. Therefore, we set $k=300$ in all our experiments. The LSI implementation that is used in our work is the Gensim python package \cite{Rehurek2010}. 

Now, we can use the English documents from the test corpus as queries. Each English document is compared by AR-LSI or CL-LSI to every Arabic document in the Arabic test corpus, and the n-best list is retrieved. This procedure is described in algorithms \ref{alg:ar_lsi} and \ref{alg:cl_lsi}.

Algorithms \ref{alg:ar_lsi} and \ref{alg:cl_lsi} describe the method to retrieve the most similar Arabic document $a_j$ to an English document $e_i$ using AR-LSI and CL-LSI respectively. Algorithm \ref{alg:ar_lsi} takes the English test corpus $C_e$ and the Arabic test corpus $C_a$. All  Arabic documents of $C_a$ are projected into AR-LSI (built from the Arabic training corpus) space. Then, each English document $e_i$ is translated into Arabic using Google MT service\footnote{\url{http://translate.google.com}}. Next, the translated document $a_{e_i}$ is projected into AR-LSI space. Then the most similar Arabic documents are retrieved from Arabic corpus. 

\begin{center}
\begin{minipage}[t]{.5\textwidth}
\vspace{0pt}  
\begin{small}
\IncMargin{2em}
\begin{algorithm}[H]
\SetKwFunction{trans}{trans}
\SetKwFunction{retrieve}{retrieve}
\SetKwFunction{evaluate}{evaluate}
\KwIn{

$C_e$: English corpus

$C_a$: Arabic corpus

$n$: number of docs to retrieve
}

$C'_e \gets \varnothing$;\ $C'_a \gets \varnothing$;\

\ForEach{doc $a_j$ in $C_a$}{
$a'_{j,\mathcal{A}} \gets a_{j,\mathcal{A}}^tU_{\mathcal{A}}S_{\mathcal{A}}^{-1}$ \tcp{map $a_j$ into AR-LSI}

put $a'_{j,\mathcal{A}}$ in $C'_a$ 
}

\ForEach{doc $e_i$ in $C_e$}{
\HiLi $a_{e_i} \gets$ \trans{$e_i$} \tcp{translate $e_i$ into Arabic}
\HiLi $a'_{e_i,\mathcal{A}} \gets a_{e_i,\mathcal{A}}^tU_{\mathcal{A}}S_{\mathcal{A}}^{-1}$ \tcp{map $a_{e_i}$ into AR-LSI}

\tcp{retrieve top-n similar documents to $e'_i$ from $C'_a$}
\HiLi $R \gets$ \retrieve{$a'_{e_i,\mathcal{A}}$, $C'_a$, $n$} 

\evaluate{$R$}  \tcp{check if $a'_i$ is in $R$}
}

\caption{{\scriptsize Retrieving Arabic documents using AR-LSI}}
\label{alg:ar_lsi}
\end{algorithm}
\DecMargin{2em}
\end{small}
\end{minipage}%
\begin{minipage}[t]{.5\textwidth}
\vspace{0pt}
\begin{small}
\IncMargin{2em}
\begin{algorithm}[H]
\SetKwFunction{trans}{trans}
\SetKwFunction{retrieve}{retrieve}
\SetKwFunction{evaluate}{evaluate}
\KwIn{

$C_e$: English corpus

$C_a$: Arabic corpus

$n$: number of docs to retrieve
}

$C'_e \gets \varnothing$;\ $C'_a \gets \varnothing$;\

\ForEach{doc $a_j$ in $C_a$}{
$a'_{j,\mathcal{C}} \gets a_{j,\mathcal{C}}^tU_{\mathcal{C}}S_{\mathcal{C}}^{-1}$ \tcp{map $a_j$ into CL-LSI}

put $a'_{j,\mathcal{C}}$ in $C'_a$ 
}

\ForEach{doc $e_i$ in $C_e$}{
\HiLi $e'_{i,\mathcal{C}} \gets e_{i,\mathcal{C}}^tU_{\mathcal{C}}S_{\mathcal{C}}^{-1}$ \tcp{map $e_i$ into CL-LSI}

\tcp{retrieve top-n similar documents to $e'_{i,\mathcal{C}}$ from $C'_a$}
\HiLi $R \gets$ \retrieve{$e'_{i,\mathcal{C}}$, $C'_a$, $n$} 

\evaluate{$R$} \tcp{check if $a'_{i,\mathcal{C}}$ is in $R$}
}

\vspace{0.80cm}

\caption{{\scriptsize Retrieving Arabic documents using CL-LSI}}
\label{alg:cl_lsi}
\end{algorithm}
\DecMargin{2em}
\end{small}
\end{minipage}
\end{center}

Retrieving Arabic documents using CL-LSI is done in the same way as AR-LSI, but machine translation service is not used. Algorithm \ref{alg:cl_lsi} describes how CL-LSI is used to retrieve Arabic documents that are comparable to an English document. Algorithm \ref{alg:cl_lsi} also takes the English test corpus $C_e$ and the Arabic test corpus $C_a$.  All documents in $C_e$ and $C_a$ are transformed into CL-LSI (built from the English-Arabic training corpus) space. Each $e_i$ is used as a query to retrieve the target pair from $C_a$ using \textit{retrieve} procedure (see below).

The difference between Algorithm \ref{alg:ar_lsi} and \ref{alg:cl_lsi} is the highlighted lines in the both algorithms. The English document in Algorithm \ref{alg:ar_lsi} is translated into Arabic first, then it is mapped to LSI space, while the English document in Algorithm \ref{alg:cl_lsi} is mapped into the LSI space directly. Machine translation is needed in Algorithm \ref{alg:ar_lsi} because the LSI model is monolingual, but machine translation is not needed in Algorithm \ref{alg:cl_lsi} because the LSI model is cross-lingual.

The \textit{retrieve} function takes the source document $d_s$, the target corpus $C_t$, and the number of documents to retrieve ($n$). The source document $d_s$ is compared with all documents in the target corpus $C_t$. The procedure then returns the top $n$ most similar documents.

\begin{procedure}[!htb]
\SetKwFunction{sort}{sort}
\KwIn{$d_{s_i}$: source document, $C_t$: target corpus, $n$: number of documents to retrieve}

$R \gets \varnothing$;\ \tcp{a list of retrieved docs}

\tcp{compute the similarity to all target docs}
\ForEach{doc $d_{t_j}$ in $C_t$}{ 
$sim \gets \cos(d_{s_i},d_{t_j})$\; put ($j$, $sim$) in $R$\;} 
\sort{$R$} \tcp {sort $R$ in descending order according to $sim$ values}

\KwRet{top $n$ elements of $R$};\

\caption{retrieve($d_{s_i}$, $C_t$, $n$)}
\label{proc:retrieve}
\end{procedure}

\subsection{Evaluation}

The evaluation of cross-lingual similarity measures (Dic-bin, Dict-cos, AR-LSI and CL-LSI) is done as follows: given $e_i$, we build the n-best list of Arabic documents (according to the similarity measure), and we check if the Arabic $a_i$ (e.g the Arabic documebt corresponding to $e_i$)  is in this list. We check this presence in the top-1 list (recall at 1 or $R@1$) and in the top-5 (recall at 5 or $R@5$). The performance measure is defined as the percentage of $a_i$, which are correctly retrieved in $R@1$ and $R@5$ lists, among all $e_i$.

\subsection{Results of text pairs retrieval (parallel)}
\label{sec:pair_retrieval}

In this experiment, we select a random sample of 100 English-Arabic sentences from the test part of each parallel corpus, which are described in Section \ref{sec:parallel_corpus}.

Each source text (English) is used as a query to retrieve exactly one relevant target text (Arabic). The experiment is conducted at the sentence level. In other words, the source sentence is used as a query to retrieve its translation in the target language. 

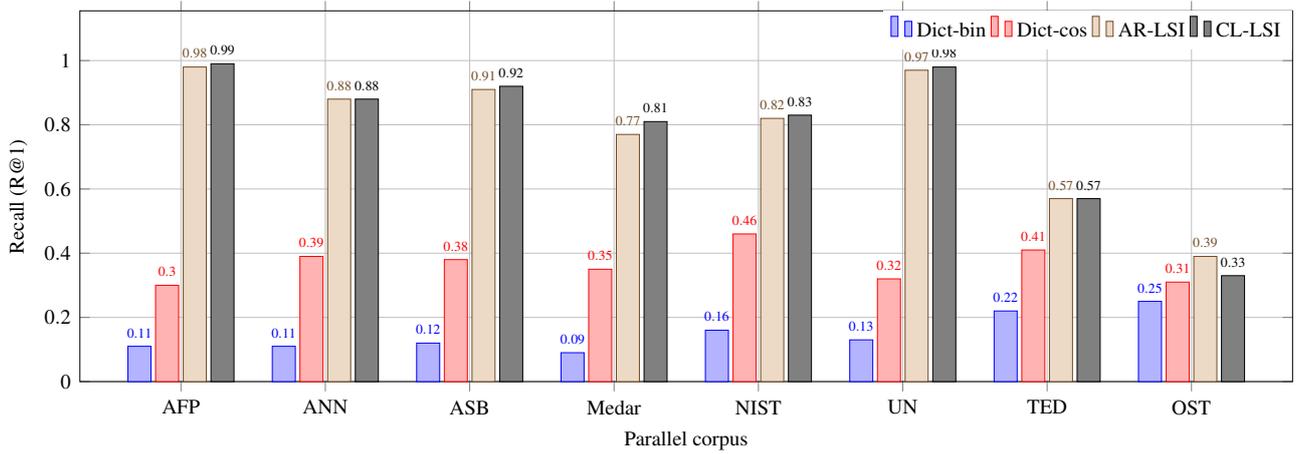
\begin{figure}[!htb]
\centering
\begin{adjustbox}{max totalsize={1.0\textwidth}{1.0\textheight},center}
\begin{tikzpicture}
\begin{axis}
[
	ybar, ymin=0, ymax=1.05,
	enlarge y limits={upper, value=0.1},
	legend style={at={(1,1)},anchor=north east,draw=none,font=\footnotesize, legend columns=0},
	ylabel={Recall (R@1)},
	xlabel={Parallel corpus},
	every axis/.append style={font=\footnotesize},
    symbolic x coords={AFP, ANN, ASB, Medar, NIST, UN,TED, OST},
    tick label style={font=\footnotesize},
    xtick=data,
    x=2.2cm,
    grid=major,
    bar width=10pt,
    nodes near coords, 
	every node near coord/.append style={font=\tiny,/pgf/number format/fixed, /pgf/number format/precision=2},
	nodes near coords align={vertical}, 
]

\addplot coordinates {
(AFP, 0.11)
(ANN, 0.11)
(ASB, 0.12)
(Medar, 0.09)
(NIST, 0.16)
(UN, 0.13)
(TED, 0.22)
(OST, 0.25)
};

\addplot coordinates {
(AFP, 0.30)
(ANN, 0.39)
(ASB, 0.38)
(Medar, 0.35)
(NIST, 0.46)
(UN, 0.32)
(TED, 0.41)
(OST, 0.31)
};

\addplot coordinates {
(AFP, 0.98)
(ANN, 0.88)
(ASB, 0.91)
(Medar, 0.77)
(NIST, 0.82)
(UN, 0.97)
(TED, 0.57)
(OST, 0.39)
};

\addplot coordinates {
(AFP, 0.99)
(ANN, 0.88)
(ASB, 0.92)
(Medar, 0.81)
(NIST, 0.83)
(UN, 0.98)
(TED, 0.57)
(OST, 0.33)
};

\legend{Dict-bin,Dict-cos,AR-LSI,CL-LSI}    
\end{axis}
\end{tikzpicture}
\end{adjustbox}
\caption{Recall ($R@1$) of retrieving parallel documents using Dict-bin, Dict-cos, AR-LSI and CL-LSI methods}
\label{fig:recall_r1}
\end{figure}

\begin{figure}[!htb]
\centering
\begin{adjustbox}{max totalsize={1.0\textwidth}{1.0\textheight},center}
\begin{tikzpicture}
\begin{axis}
[
	ybar, ymin=0, ymax=1.05,
	enlarge y limits={upper, value=0.1},
	legend style={at={(1,1)},anchor=north east,draw=none,font=\footnotesize, legend columns=0},
	ylabel={Recall (R@5)},
	xlabel={Parallel corpus},
	every axis/.append style={font=\footnotesize},
    symbolic x coords={AFP, ANN, ASB, Medar, NIST, UN,TED, OST},
    tick label style={font=\footnotesize},
    xtick=data,
    x=2.2cm,
    grid=major,
    bar width=10pt,
    nodes near coords, 
	every node near coord/.append style={font=\tiny,/pgf/number format/fixed, /pgf/number format/precision=2},
	nodes near coords align={vertical}, 
]

 \addplot coordinates {
(AFP, 0.24)
(ANN, 0.24)
(ASB, 0.32)
(Medar, 0.22)
(NIST, 0.32)
(UN, 0.30)
(TED, 0.45)
(OST, 0.43)
};

\addplot coordinates {
(AFP, 0.69)
(ANN, 0.72)
(ASB, 0.72)
(Medar, 0.71)
(NIST, 0.78)
(UN, 0.63)
(TED, 0.76)
(OST, 0.55)
};

\addplot coordinates {
(AFP, 1.00)
(ANN, 0.96)
(ASB, 0.96)
(Medar, 0.92)
(NIST, 0.94)
(UN, 1.00)
(TED, 0.74)
(OST, 0.61)
};

\addplot coordinates {
(AFP, 1.00)
(ANN, 0.96)
(ASB, 0.98)
(Medar, 0.97)
(NIST, 0.91)
(UN, 1.00)
(TED, 0.82)
(OST, 0.76)
};

\legend{Dict-bin,Dict-cos,AR-LSI,CL-LSI}    
\end{axis}
\end{tikzpicture}
\end{adjustbox}
\caption{Recall ($R@5$) of retrieving parallel documents using Dict-bin, Dict-cos, AR-LSI and CL-LSI methods}
\label{fig:recall_r5}
\end{figure}
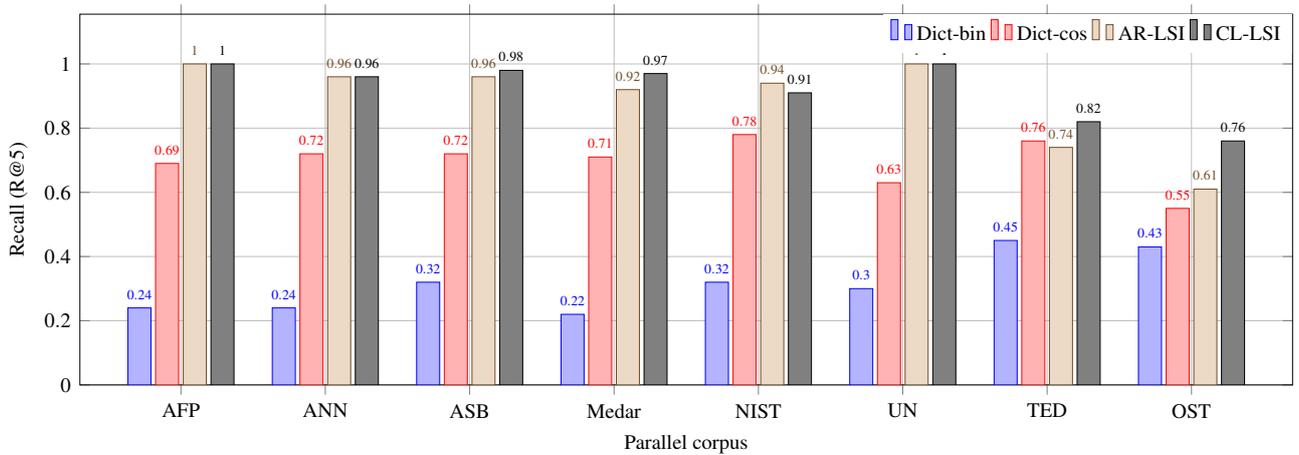

Figures \ref{fig:recall_r1} and \ref{fig:recall_r5} present the results of the first and the fifth recall for parallel corpora, using Dic-bin, Dict-cos, AR-LSI and CL-LSI methods. Dict-bin is computed using Formula \ref{eq:binCM_sym} and Dic-cos is computed using Formula \ref{eq:cosine}. The figures show that both LSI methods (AR-LSI and CL-LSI) have better recall than the dictionary based methods (Dict-bin and Dict-cos). It can be concluded that both LSI methods are better and more robust than the dictionary based methods since it does not need any dictionary or morphological analysis, and it is language independent.

Comparing Dict-bin with Dict-cos, the results show that cosine measure achieves better results than the binary measure in terms of recall scores. However, the recall scores for dictionary based measures are still limited. This is due to the limitations of the dictionary and the morphological tools. Besides that, word-to-word translations based on dictionaries can lead to many errors (translation ambiguity). 

In general, the recall of AR-LSI and CL-LSI methods for AFP, ANN, ASB, Medar, NIST and UN corpora is better than the one for TED and OST corpora. This is maybe because the latter corpora are generated by volunteer translators, while the former corpora are generated by professional translators.

Comparing AR-LSI and CL-LSI methods, it is not easy to get a general conclusion about the performance of LSI since it depends on the nature of the corpus and on the desired recall ($R@1$ or $R@5$). For example,  for AFP, ASB, NIST, Medar, and UN corpora, CL-LSI is slightly better than AR-LSI for $R@1$. In contrast, for OST, AR-LSI is better than CL-LSI. The performance of the CL-LSI is equal to, or better than the AR-LSI in 6 out of 8 of corpora for $R@1$.

We checked the significance of differences of the results using McNemar's test \cite{McNemar1947}. The conclusion is that they are not significantly different. Therefore, both approaches obtain mostly similar performance. However, it should be noted that CL-LSI does not require a MT system. Therefore, we can affirm that CL-LSI is competitive compared to AR-LSI.

The results of AR-LSI show that MT can be sufficient for cross-lingual retrieval. However, to investigate the effect of the performance of the MT system on the performance of the AR-LSI, we run an experiment to simulate a perfect MT system (The Oracle experiment). This is done by retrieving an Arabic document by providing the same document as a query. In other words, source and target documents are the same. This experiment is done on all corpora and the results of $R@1$ is 1.0 for each corpus of the parallel corpora. This result reveals the lack of robustness of AR-LSI according to the MT system's performance.  

Finally, we compare our LSI result to the results in \cite{Littman1998}. The authors of the paper worked on French-English document retrieval using LSI. They applied the method on Hansard parallel corpus\footnote{\url{www.isi.edu/natural-language/download/hansard}}, which is the proceedings of the Canadian parliament. UN corpus in our work is close (in terms of domain) to Hansard corpus. Therefore, the results can be compared. \cite{Littman1998} reported 0.98 of R@1 on English-French texts of Hansard corpus and, we achieved a similar result on English-Arabic texts of the UN corpus using the CL-LSI method.

\subsection{Results of comparable document pairs retrieval}

The previous experiments on parallel corpora have been conducted in order to show the feasibility of the proposed methods. Since, the results are good for the two last methods, we will now use them on comparable corpora.

The same experimental protocol as described in Section \ref{sec:experimental_procedure} is applied to retrieve the documents of comparable corpus. The source document is used as a query to retrieve its pair in the target language. The difference is that the CL-LSI matrix is built using the training part of the comparable corpus. The objective of this experiment is to investigate the use of comparable corpora for training CL-LSI in order to retrieve cross-lingual documents. In this experiment, the source document (English) is used as a query to retrieve its target comparable document (exactly one relevant Arabic document). 

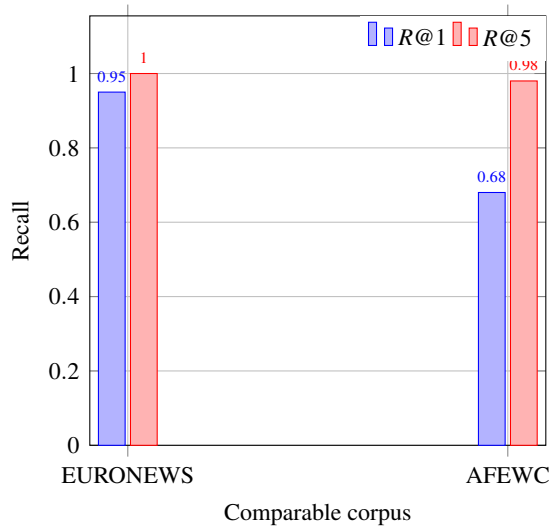
\begin{figure}[!htb]
\centering
\begin{tikzpicture}
\begin{axis}
[
	ybar, ymin=0, ymax=1.05,
	enlarge y limits={upper, value=0.1},
	legend style={at={(1,1)},anchor=north east,draw=none,font=\footnotesize, legend columns=0},
	ylabel={Recall},
	xlabel={Comparable corpus},
	every axis/.append style={font=\footnotesize},
    symbolic x coords={EURONEWS, AFEWC},
    tick label style={font=\footnotesize},
    xtick=data,
    x=5cm,
    grid=major,
    bar width=10pt,
    nodes near coords, 
	every node near coord/.append style={font=\tiny,/pgf/number format/fixed, /pgf/number format/precision=2},
	nodes near coords align={vertical}, 
]

\addplot coordinates {
(EURONEWS, 0.95)
(AFEWC, 0.68)

};

\addplot coordinates {
(EURONEWS, 1.00)
(AFEWC, 0.98)

};
    
\legend{$R@1$,$R@5$}    
\end{axis}
\end{tikzpicture}
\caption{Recall of retrieving comparable documents using CL-LSI method}
\label{fig:lsi_recall_comparable}
\end{figure}

Results of retrieving comparable documents (at the document level) using CL-LSI  are presented in Figure \ref{fig:lsi_recall_comparable}. The figure shows the recall scores of the CL-LSI method on EURONEWS and AFEWC comparable corpora. The recall of CL-LSI on EURONEWS corpus is better than on AFEWC corpus. This could be due to the fact that EURONEWS articles are mostly translations of each other, while Wikipedia articles are not necessarily translations of each other as mentioned in Section \ref{sec:comparable_corpus}.

From Figures \ref{fig:recall_r1}, \ref{fig:recall_r5} and  \ref{fig:lsi_recall_comparable}, it can be noted that CL-LSI can retrieve the target information at the document level and at the sentence level with almost the same performance.

In this section we experimented two cross-lingual similarity measures: based on bilingual dictionary and based on LSI. 

We further proposed a morphological analysis technique (\textit{MorphAr}) for Arabic words to match them with English words. We experimentally investigated different combinations of English-Arabic morphological analysis techniques to determine the best one to match English-Arabic words. We found that \textit{MorphAr} for Arabic and lemmatization for English lead together to best matching rate for English and Arabic words. We noted that the dictionary based method has limited performance certainly due to the limitations of the bilingual dictionaries and the morphological  analysis tools. Moreover, word-to-word matching based on dictionaries can lead to many errors.

We  used the LSI method in two ways: monolingual (AR-LSI) and cross-lingual (CL-LSI). The first method needs to use a machine translation system in order to translate the source into the language of the target text, while the second method merges the training data of both languages. In the test step, the comparison is done between vectors of the same type.

We applied these methods on several corpora and the results showed that the CL-LSI can be competitive to the AR-LSI. The advantage of CL-LSI is that it  does not need machine translation. The results also showed that the method can be used to retrieve comparable pairs. Both CL-LSI and AR-LSI achieved better results than the dictionary based method. In addition, LSI methods do not need morphological analysis tools or bilingual dictionaries, and it overcomes the problem of vocabulary mismatch between queries, and they are language independent.

Since we concluded that CL-LSI is better than the dictionary based measure, we use CL-LSI in the next section to align cross-lingual documents collected from other different sources.

\section{Aligning Comparable Documents Collected From Different Sources}
\label{sec:cross_lingual_alignment}

As we mentioned in the introduction, comparable documents can be interesting and very useful for many applications such as comparing news articles or product reviews. i.e., the journalist may be interested in what is being said about an event in local and foreign media. In addition to that, aligning comparable documents enriches language resources for low-resourced languages. Therefore, we present in this section an alignment method that aligns comparable documents collected from other sources than Wikipedia and EURONEWS. In this section, we focus on aligning English-Arabic news articles collected from the Internet.  The English news are collected from the British Broadcast Corporation (BBC) website\footnote{www.bbc.com/news}, and the Arabic news are collected from ALJAZEERA (JSC) website\footnote{www.aljazeera.net}. 

We use the CL-LSI method that is presented in Section \ref{sec:cl_lsi_method} to align cross-lingual news articles. The task is to align English-Arabic news articles that are related to the same news story or event. In other words, for a given source document, the objective is to retrieve and align the most relevant target document (same news story) and not all similar documents. For example, if the English document is related to ``elections in France", we want to retrieve the Arabic document that is related to the same news story, and not any other news article related to ``elections". Therefore, this task is more challenging compared to the work in the previous section. This section inspects the ability of CL-LSI to perform automatic alignment at the event level. In the previous section we used CL-LSI for document pairs retrieval, and the corpora were already aligned, but in this section, we use CL-LSI to align comparable documents, which are collected from different sources (BBC-JSC news articles), and they are not already aligned.

\subsection{The Proposed Method}

The CL-LSI approach needs a parallel or comparable corpus for training as described in Section \ref{sec:cl_lsi_method}. To build the CL-LSI $\mathcal{C}$ matrix, we use EURONEWS comparable corpus, which is described in Section \ref{sec:comparable_corpus}. Then, we apply LSI to obtain $U_{\mathcal{C}}S_{\mathcal{C}}V_{\mathcal{C}}^T$. All text documents are preprocessed by removing punctuation marks, stop words (common words), and words that occurred less than three times (low-frequency words) in the corpus.

\begin{figure}[!htb]
\centering
\begin{adjustbox}{max totalsize={1.0\textwidth}{1.0\textheight},center}
\begin{tikzpicture}
\tikzstyle{every node}=[thick]
\tikzstyle{every path}=[thick]
\tikzstyle{every draw}=[thick]

\node [corpus, align=center, minimum height=1cm] (bbc) {\\BBC\\2012-2013};
\node [corpus, align=center, below of=bbc, yshift=-1.5cm, font=\small] (bbc_jan) {\\\\Jan.\\2012};
\node [corpus, align=center, below of=bbc_jan, yshift=-0.8cm, font=\small] (bbc_feb) {\\\\Feb.\\2012};
\node [corpus, align=center, below of=bbc_feb, yshift=-1.5cm, font=\small] (bbc_nov) {\\\\Nov.\\2013};
\node [corpus, align=center, below of=bbc_nov, yshift=-0.8cm, font=\small] (bbc_dec) {\\\\Dec.\\2013};

\draw [thick, densely dotted] (bbc_feb) -- (bbc_nov);

\node [corpus, align=center, minimum height=1cm, left of=bbc, xshift=7cm] (jsc) {\\JSC\\2012-2013};
\node [corpus, align=center, below of=jsc, yshift=-1.5cm, font=\small] (jsc_jan) {\\\\Jan.\\2012};
\node [corpus, align=center, below of=jsc_jan, yshift=-0.8cm, font=\small] (jsc_feb) {\\\\Feb.\\2012};
\node [corpus, align=center, below of=jsc_feb, yshift=-1.5cm, font=\small] (jsc_nov) {\\\\Nov.\\2013};
\node [corpus, align=center, below of=jsc_nov, yshift=-0.8cm, font=\small] (jsc_dec) {\\\\Dec.\\2013};

\draw [thick, densely dotted] (jsc_feb) -- (jsc_nov);

\node [rec, align=center, below of=bbc, yshift=-5cm, xshift=3cm] (lsi) {CL-LSI\\trained on\\EURONEWS\\corpus};

\node [corpus, align=center, below of=lsi, yshift=-2.5cm] (bbc_jsc) {Top-n\\BBC-JSC\\aligned articles};

\path [arrow] (bbc) -- (bbc_jan) node [midway, anchor=east, font=\small] {split by month};

\path [arrow] (jsc) -- (jsc_jan) node [midway, anchor=west, font=\small] {split by month};

\path [arrow,thick] (lsi) -- (bbc_jsc) node [midway, anchor=east, font=\small] {align};

\draw (bbc_jan) edge[thick,draw=blue,out=0,in=90,->]  (lsi) ;
\draw (jsc_jan) edge[thick,draw=blue,out=180,in=90,->]  (lsi) ;

\draw (bbc_feb) edge[thick,draw=red,out=0,in=135,->]  (lsi) ;
\draw (jsc_feb) edge[thick,draw=red,out=180,in=45,->]  (lsi) ;

\draw (bbc_nov) edge[thick,draw=orange,out=0,in=180,->]  (lsi) ;
\draw (jsc_nov) edge[thick,draw=orange,out=180,in=0,->]  (lsi) ;

\draw (bbc_dec) edge[thick,draw=cyan,out=0,in=200,->]  (lsi) ;
\draw (jsc_dec) edge[thick,draw=cyan,out=180,in=340,->]  (lsi) ;

\end{tikzpicture}
\end{adjustbox}
\caption{Automatic alignment of BBC and JSC news stories}
\label{fig:automatic_aligning}
\end{figure}
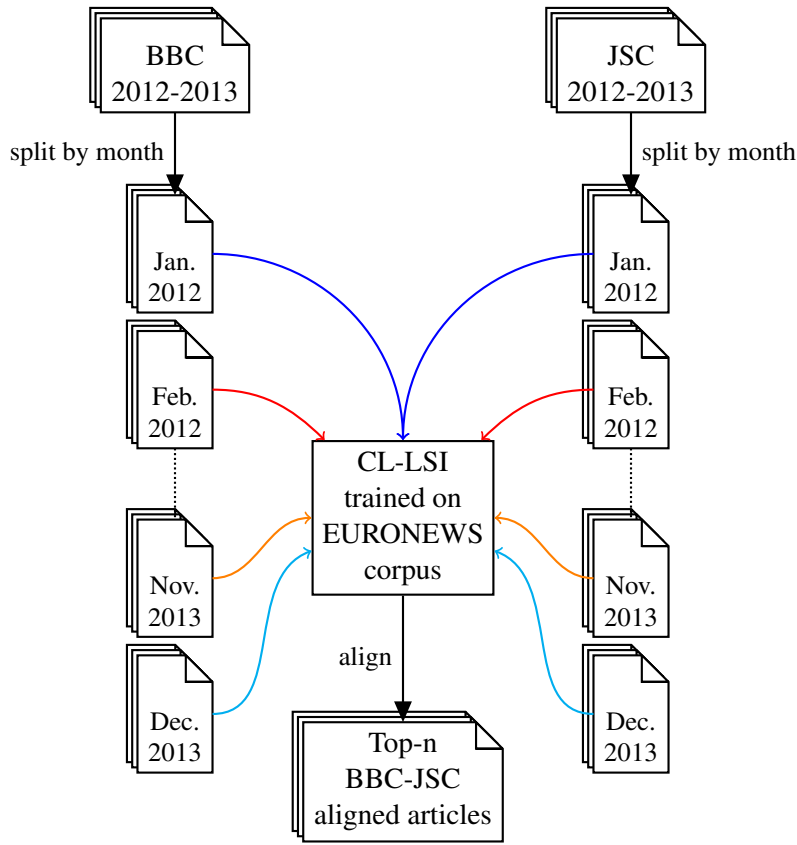

As mentioned earlier, the objective is to align English articles, which are collected from BBC news website, with Arabic news articles, which are collected from JSC website. First, we crawl BBC and JSC websites to collect news articles published in 2012 and 2013 using httrack tool\footnote{www.httrack.com}. The articles of the BBC-JSC corpus are then split into several sub-corpora. Each sub-corpus is composed of news articles that are published in a specific month. Consequently, we obtain 24 sub-corpora for each language as shown in Figure \ref{fig:automatic_aligning}. The number of articles in each month-corpus ranges between 70 and 300.

\begin{center}
\begin{minipage}[t]{.5\textwidth}
\vspace{0pt}  
\begin{small}
\IncMargin{2em}
\begin{algorithm}[H]
\SetKwFunction{align}{align}
\SetKwFunction{sort}{sort}
\KwIn{$C_e$: English corpus, $C_a$: Arabic corpus}
\KwResult{top $N$ aligned articles}
$L \gets \varnothing$;\ $C'_e \gets \varnothing$;\ $C'_a\gets \varnothing$;\

\tcp{map $e_i$ and $a_j$ into CL-LSI}

\ForEach{document $e_i$ in $C_e$}{
$e'_i \gets e_i^tU_{\mathcal{C}}S_{\mathcal{C}}^{-1}$;\ put $e'_i$ in $C'_e$;\
}
\ForEach{document $a_j$ in $C_a$}{
$a'_j \gets a_j^tU_{\mathcal{C}}S_{\mathcal{C}}^{-1}$;\ put $a'_j$ in $C'_a$;\
}

\tcp{align the most similar $a_j$ to each $e_i$ }

\ForEach{document $e'_i$ in $C'_e$}{
($a_j$, $sim$) $\gets$ \align{$e'_i$, $C'_a$}\; put ($e_i$, $a_j$, $sim$) in $L$\;
}
\sort{$L$} \tcp{sort descendingly}
Select top $N$ elements from $L$;\
\caption{{\tiny Aligning English-Arabic docs}}
\label{alg:aligning}
\end{algorithm}
\DecMargin{2em}
\end{small}
\end{minipage}%
\begin{minipage}[t]{.5\textwidth}
\vspace{0pt}
\begin{small}
\IncMargin{2em}
\begin{procedure}[H]
\KwIn{$e'_i$, $C'_a$}
\KwOut{$a_j$, $sim$}

$L \gets \varnothing$;\ \tcp{a list of candidate Arabic document}
$sim_{max} \gets 0$\;

\ForEach{document $a'_j$ in $C'_a$}{ 
\tcp{compare $a'_j$ to all documents in $C'_a$}
$sim \gets \cos(e'_i,a'_j)$;\ 

\If{$sim > sim_{max}$} {
    $sim_{max} \gets sim$\;
    $a \gets a_j$;\
  }
}
\KwRet{$a$, $sim_{max}$};\

\vspace{1.2 cm}

\caption{align($e'_i$, $C'_a$)}
\label{proc:aliging_function}
\end{procedure}
\DecMargin{2em}
\end{small}

\end{minipage}
\end{center}

Each BBC sub-corpus and its corresponding JSC sub-corpus are provided to the CL-LSI to perform the automatic alignment as shown in Figure \ref{fig:automatic_aligning}. 

The alignment steps are described in Algorithm \ref{alg:aligning}. The approach we propose aligns English and Arabic documents of a month-corpus. The process is repeated for each month. To align an English document ($e_i$) and an Arabic document ($a_j$) of a month-corpus, first the English $C_e$ month-corpus and the Arabic $C_a$ month-corpus are provided to the algorithm which maps English $e_i$ and Arabic $a_i$ into the CL-LSI space. Then, the algorithm aligns each English document to the most relevant Arabic documents. The aligned articles with their similarity value $(a_i,e_i,sim)$ are added to the list $L$,  which is sorted later in descending order according to the similarity value.

The $align$ procedure which is called in the algorithm takes the English document $e'_i$ and the Arabic corpus $C'_a$. Then, the procedure computes the similarity between $e'_i$ and all $a'_j$ of $C'_a$ and returns $a'_j$ that has the highest similarity value.

The output of Algorithm \ref{alg:aligning} is a list of top-n most similar document pairs. If the aligned document pairs are related to the same story (checked manually), then they are considered to be correctly aligned. Otherwise, they are considered to be misaligned. The list of top-n most similar document pairs, is checked by hand to make sure that document pairs are correctly aligned. We remind that the objective of the experiment is to align news articles such that they are related to the same news story or event, and not to retrieve the articles sharing the same generic topic. This handwork is done on the top-15 article pairs retrieved from each month-corpus. The total number of documents to be validated is 360 article pairs. In the next section we present the results of our method.

\subsection{Results}

\begin{figure}[!htb]
\begin{adjustbox}{max totalsize={1.0\textwidth}{1.0\textheight},center}
\begin{tikzpicture}
\begin{axis}
[
	ybar, ymin=0, 
	ylabel={Accuracy (\%)},
	xlabel={Month/year},
    symbolic x coords={Jan 2012, Feb 2012, Mar 2012, Apr 2012, May 2012, Jun 2012, Jul 2012, Aug 2012, Sep 2012, Oct 2012, Nov 2012, Dec 2012, Jan 2013, Feb 2013, Mar 2013, Apr 2013, May 2013, Jun 2013, Jul 2013, Aug 2013, Sep 2013, Oct 2013, Nov 2013, Dec 2013},
    y tick label style={font=\scriptsize},
    xtick=data,
    x=0.55cm,
    x tick label style={rotate=90, font=\scriptsize},
    grid=major,
    bar width=12pt,
    nodes near coords, 
	every node near coord/.append style={font=\tiny},
	nodes near coords align={vertical}, 
]

\addplot coordinates {
(Jan 2012,  0.73)
(Feb 2012,  0.87)
(Mar 2012,  0.93)
(Apr 2012,  0.87)
(May 2012,  0.73)
(Jun 2012,  0.80)
(Jul 2012,  0.80)
(Aug 2012,  0.80)
(Sep 2012,  0.87)
(Oct 2012,  1.00)        
(Nov 2012,  1.00)
(Dec 2012,  1.00)

(Jan 2013,  0.87)
(Feb 2013,  0.80)
(Mar 2013,  1.00)
(Apr 2013,  0.73)
(May 2013,  0.87)
(Jun 2013,  0.87)
(Jul 2013,  0.80)
(Aug 2013,  0.73)
(Sep 2013,  0.87)
(Oct 2013, 0.87)        
(Nov 2013, 0.80)
(Dec 2013, 0.73)
};

\end{axis}
\end{tikzpicture}
\end{adjustbox}
\caption{Accuracy of articles alignment for years 2012 and 2013}
\label{fig:alignment_accuracy}
\end{figure}
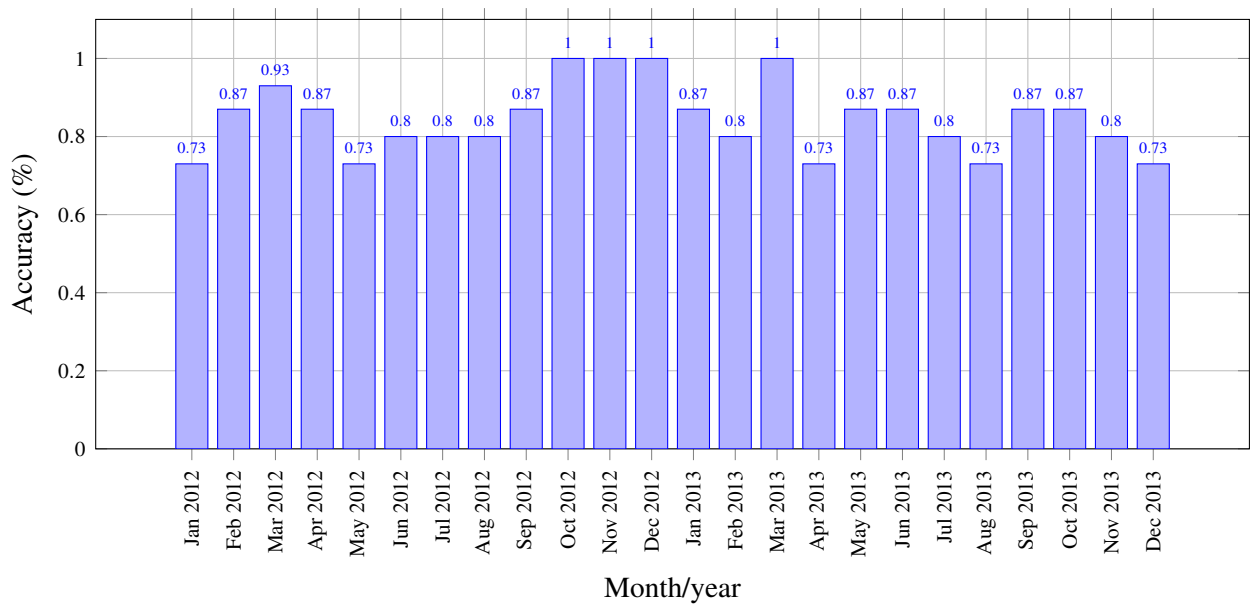

Experimental results are presented in Figure \ref{fig:alignment_accuracy}. The figure shows the accuracy of alignment of the top-15 most similar documents of each month of the 24 month-corpus corresponding to the years 2012--2013. The accuracy of the alignment is defined as the number of cross-lingual articles, that are correctly aligned, divided by the total number of articles.

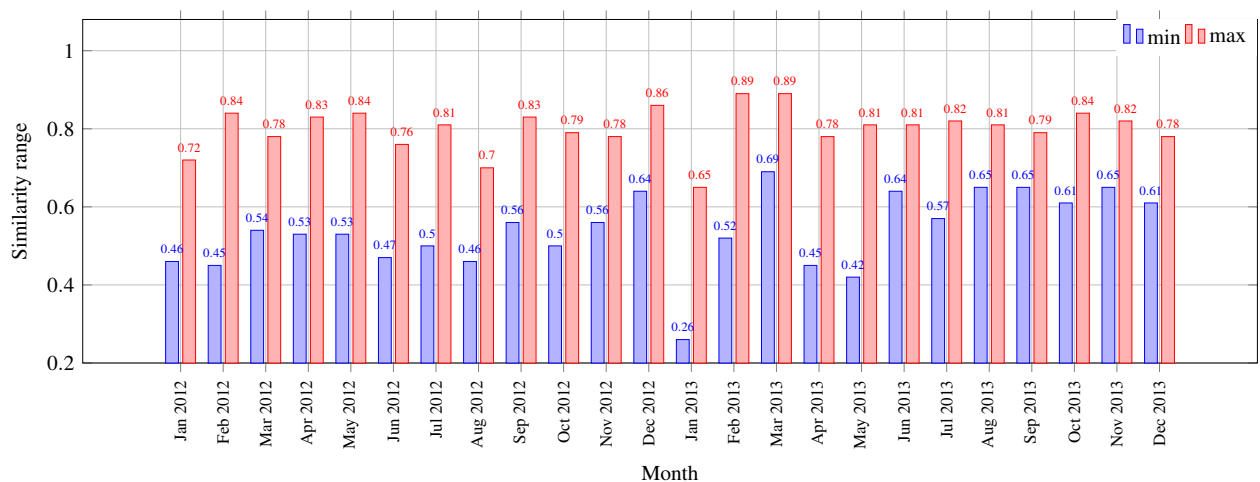
\begin{figure}[!htb]
\centering 
\begin{tikzpicture}[scale=0.8]
\tikzstyle{every node}=[font=\small]
\begin{axis}
[
	ybar,
    enlarge y limits={upper, value=0.1},
	x=0.7cm,
    bar width=6pt,
	ymax = 1.0,	
	ymin = 0.2,
	xlabel={Month},
	xlabel style={name=xlabel},
	ylabel={Similarity range},
	xtick=data,
    nodes near coords, 
	every node near coord/.append style={font=\tiny,/pgf/number format/fixed, /pgf/number format/precision=2},
	nodes near coords align={vertical}, 
	grid=major,
	y tick label style={font=\small},
	x tick label style={rotate=90, font=\scriptsize},
    legend style={at={(1,1)},anchor=north east,draw=none,font=\small, legend columns=0},
    symbolic x coords={Jan 2012, Feb 2012, Mar 2012, Apr 2012, May 2012, Jun 2012, Jul 2012, Aug 2012, Sep 2012, Oct 2012, Nov 2012, Dec 2012, Jan 2013, Feb 2013, Mar 2013, Apr 2013, May 2013, Jun 2013, Jul 2013, Aug 2013, Sep 2013, Oct 2013, Nov 2013, Dec 2013},
]

\addplot coordinates {
	(Jan 2012,	0.46 )    
    (Feb 2012,	0.45 )   
    (Mar 2012,	0.54 )
	(Apr 2012,	0.53 )  
    (May 2012,	0.53 )  
    (Jun 2012,	0.47 )
	(Jul 2012,	0.50 ) 
    (Aug 2012,	0.46 ) 
    (Sep 2012,	0.56 )
    (Oct 2012, 	0.50 )
    (Nov 2012,	0.56 )
    (Dec 2012,	0.64 )  
    (Jan 2013,	0.26 )    
    (Feb 2013,	0.52 )   
    (Mar 2013,	0.69 )
	(Apr 2013,	0.45 )  
    (May 2013,	0.42 )  
    (Jun 2013,	0.64 )
	(Jul 2013,	0.57 ) 
    (Aug 2013,	0.65 ) 
    (Sep 2013,	0.65 )
    (Oct 2013, 	0.61 )
    (Nov 2013,	0.65 )
    (Dec 2013,	0.61 )
};

\addplot coordinates {
	(Jan 2012,	0.72 )    
    (Feb 2012,	0.84 )   
    (Mar 2012,	0.78 )
	(Apr 2012,	0.83 )  
    (May 2012,	0.84 )  
    (Jun 2012,	0.76 )
	(Jul 2012,	0.81 ) 
    (Aug 2012,	0.70 ) 
    (Sep 2012,	0.83 )
    (Oct 2012, 	0.79 )
    (Nov 2012,	0.78 )
    (Dec 2012,	0.86 )   
   	(Jan 2013,	0.65 )    
    (Feb 2013,	0.89 )   
    (Mar 2013,	0.89 )
	(Apr 2013,	0.78 )  
    (May 2013,	0.81 )  
    (Jun 2013,	0.81 )
	(Jul 2013,	0.82 ) 
    (Aug 2013,	0.81 ) 
    (Sep 2013,	0.79 )
    (Oct 2013, 	0.84 )
    (Nov 2013,	0.82 )
    (Dec 2013,	0.78 )
};

\legend{min,max}
\end{axis}
\end{tikzpicture}
\caption{Similarity ranges of the top-15 similar documents of BBC-JSC of the years 2012 and 2013}
\label{fig:sim_range}
\end{figure}

The ranges of similarity values of the top-15 aligned articles for the years 2012 and 2013 are shown in Figure \ref{fig:sim_range}. The figure shows the minimum and the maximum of similarity values for each month. For 2012, the maximum value is 0.86 and the minimum is 0.45. For 2013, the maximum value is 0.89 and the minimum is 0.26. It can be noted from the figure that the similarity ranges (minimum and maximum values) are close to each others for all months in 2012 and 2013 except for Jan., Feb., Apr. and May 2013. This is maybe due to the nature of crawled articles for each month, where the crawling tool may miss some articles in the crawling process. Furthermore, Figure \ref{fig:sim_range} shows that min-max values vary from a month to another. This is why we decide to choose the top-N similar articles rather than setting a threshold for the similarity value.

The accuracy of correctly aligned documents is 0.85 (305 out of 360). We carried out more investigations about misaligned articles during the validation process. We found that they are all related to the same topic domain, but they are not related to the same news story or event. The investigation reveals that some of these articles are misaligned despite of high similarity. The reason is that they are related to the same event, but this event happened in different countries. For instance, one of misaligned news articles were related to the elections, but the English article was related to the elections in Bulgaria, while the Arabic article was related to elections in Pakistan. We conducted a search for ``elections in Bulgaria" in our JSC collection, but we could not find any news article that is related to elections in Bulgaria. We also found that some of these stories are local news that are covered only by either JSC or BBC. Besides that, it should also be noted that the crawling tool sometimes can not crawl all the web pages from the website. This is why for some months, some news stories could not be found either in the BBC or JSC collections.

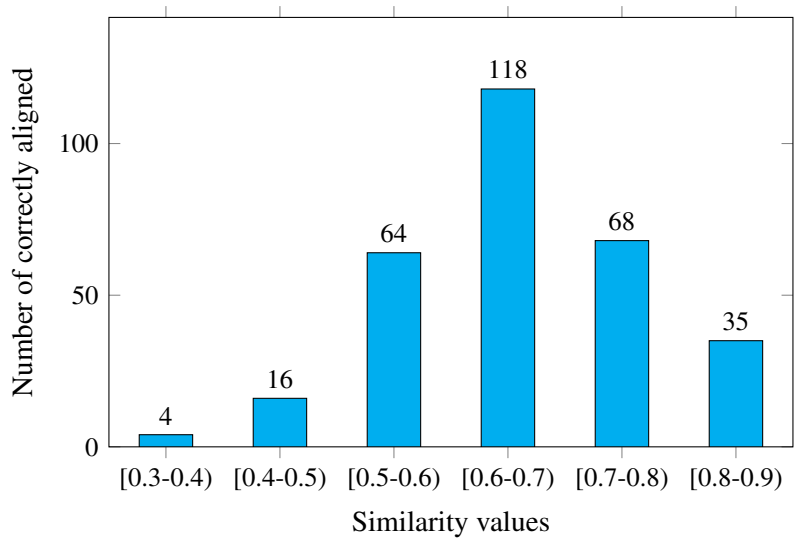
\begin{figure}[!htb]
\centering
\begin{tikzpicture}
\begin{axis}[
	ybar, ymin=0,
	enlarge y limits={upper, value=0.2},
	legend style={at={(0.5,-0.15)}, anchor=north,legend columns=0},
	ylabel={Number of correctly aligned},
	xlabel={Similarity values},
    symbolic x coords={[0.3-0.4), [0.4-0.5), [0.5-0.6), [0.6-0.7), [0.7-0.8), [0.8-0.9)},
    tick label style={font=\small},
    xtick=data,
    nodes near coords, 
    every node near coord/.style={font=\small},
    x=1.5cm,
    bar width=20pt,
  ]
    \addplot[ybar,fill=cyan] coordinates {
        ([0.3-0.4),  4)
        ([0.4-0.5),  16)
        ([0.5-0.6),  64)
        ([0.6-0.7),  118)
        ([0.7-0.8),  68)
        ([0.8-0.9),  35)        
    };
\end{axis}
\end{tikzpicture}
\caption{Similarity values vs. number of correctly aligned articles}
\label{fig:alignment_histo}
\end{figure}

Figure \ref{fig:alignment_histo} shows the number of correctly aligned articles vs. their similarity values. The similarity values in this figure  are divided into intervals. The number of correctly aligned articles increases as the similarity value increases, up to the interval $[0.6-0.7)$, then it decreases for higher similarity values. The interpretation might be as follows: when the similarity is low, the articles are mostly related to the same topic but not the same news story. As the similarity increases, the likelihood for the aligned articles to be related to the same news story increases up to a certain value, then it normally decreases again. This is because it is unlikely to find related news articles written by BBC and JSC (different news agencies), that have a high similarity value at the same time. 

At the end, we got 305 documents of the BBC-JSC corpus, for which the alignments are checked by hand. These resources are important for our study, because we can use it to study comparable news of BBC and JSC. 



In this section, we used CL-LSI to align comparable news documents collected from BBC and JSC. We showed that CL-LSI is able to not only align cross-lingual documents collected from the same source based on topics, but it can also align cross-lingual news articles collected from different sources based on events. The results showed that 85\% of cross-lingual articles are correctly aligned. Also we demonstrated that CL-LSI method can be reliable to align cross-lingual news.

\section{Conclusion and Future Work}

In this research, we provided methods for collecting, retrieving and aligning comparable documents. 

First, we collected comparable corpora from Wikipedia and EURONEWS in Arabic, English and French languages, they are all aligned at the document level. 


Then, we investigated two cross-lingual similarity measures to retrieve and align English-Arabic comparable documents. The first measure is based on  bilingual dictionary, and the second is based on Latent Semantic Indexing (LSI). The experiments on several corpora showed that the Cross-Lingual LSI (CL-LSI) measure outperformed the dictionary based measure. These conclusions are based on several corpus, parallel and comparable, and from different sources, different natures. The advantage of CL-LSI is that it needs neither bilingual dictionaries nor morphological analysis tools. Moreover, it overcomes the problem of vocabulary mismatch between documents.

Moreover, we also collected English-Arabic comparable news documents from local and foreign sources.  The English documents are collected from the British Broadcast Corporation (BBC), while the Arabic documents are collected from ALJAZEERA (JSC) news websites. 


After, we used CL-LSI to align BBC-JSC news documents. The evaluation of the alignment has shown that CL-LSI is not only able to align cross-lingual documents at the topic level, but also it is able to do this at the event level.

In future, we will collect and study comparable documents collected from other sources and in other languages, especially those of social networks. It would be interesting to show how the methods proposed in this paper perform with this special kind of data. Moreover, we will use the CL-LSI method to align the cross-lingual texts at the sentence level. These aligned sentences can be useful to train machine translation systems. We will study also the impact of different text preprocessing techniques, such as stemming, lemmatization and linguistic features, on the CL-LSI method.

\newpage

\medskip
\bibliography{myRef}

\end{document}